\documentclass[letterpaper,twocolumn,10pt]{article}
\usepackage{usenix}
 \usepackage{tabularray}
\usepackage{tikz}
\usepackage{authblk}
\usepackage{amsmath}
\usepackage{comment}
\usepackage{filecontents}

\usepackage{algorithm}
\usepackage{algorithmicx}
\usepackage{algpseudocode}
\usepackage{multirow}
\usepackage{booktabs}

\newcommand{\method}{SPEX}

\begin{document}
\title{Breaking the Reward Barrier: Accelerating Tree-of-Thought Reasoning via \underline{Sp}eculative \underline{Ex}ploration}

\author[1,2]{\rm Shuzhang Zhong}
\author[1,2]{\rm Haochen Huang}
\author[1,2]{\rm Shengxuan Qiu}
\author[3]{\rm Pengfei Zuo$^{*}$}
\author[2]{\rm Runsheng Wang}
\author[1,2]{\rm Meng Li$^{*}$}
\affil[1]{Institute for Artificial Intelligence, Peking University}
\affil[2]{School of Integrated Circuits, Peking University}
\affil[3]{ByteDance Seed}

\maketitle 
 \begin{abstract}
Tree-of-Thought (ToT) reasoning structures Large Language Model (LLM) inference as a tree-based search, demonstrating strong potential for solving complex mathematical and programming tasks. However, its efficiency is constrained by the \textit{reward dependency barrier}---a synchronization bottleneck caused by sequential reward-guided exploration that limits search parallelism and introduces substantial latency. Prior system optimizations, mainly designed for linear Chain-of-Thought (CoT) reasoning, cannot address these challenges, leaving the efficiency of ToT underexplored.

To enhance ToT reasoning efficiency, we observe that the reasoning paths can be explored speculatively to break the reward synchronization barrier. Therefore, in this paper, we propose \method~and introduce three key techniques: (i) intra-query speculative path selection to predict and expand high-potential branches of ToT, (ii) inter-query budget allocation to balance speculative resource allocation across queries dynamically, and (iii) adaptive early termination to prune deep and redundant branches for a skewed search tree.



We implement SPEX on top of the SGLang framework and evaluate it across diverse ToT algorithms and LLMs. Extensive experiments show that SPEX achieves $1.2 \sim 3 \times$ speedup for different ToT reasoning algorithms. Moreover, SPEX synergizes with token-level speculative decoding, achieving cumulative speedups of up to $4.1\times$. Ablation studies further confirm the contributions of each technique. Overall, SPEX represents a significant step toward efficient and scalable ToT reasoning, unlocking the parallelism required for high-performance inference-time scaling for LLMs.
\end{abstract}

\begingroup
\renewcommand\thefootnote{}
\footnotetext{$^{*}$ Corresponding authors: 
Meng Li (meng.li@pku.edu.cn), 
Pengfei Zuo (pfzuo.cs@gmail.com)}
\endgroup

\section{Introduction}


\begin{figure}[!tb]
    \centering
    \includegraphics[width=0.95\linewidth]{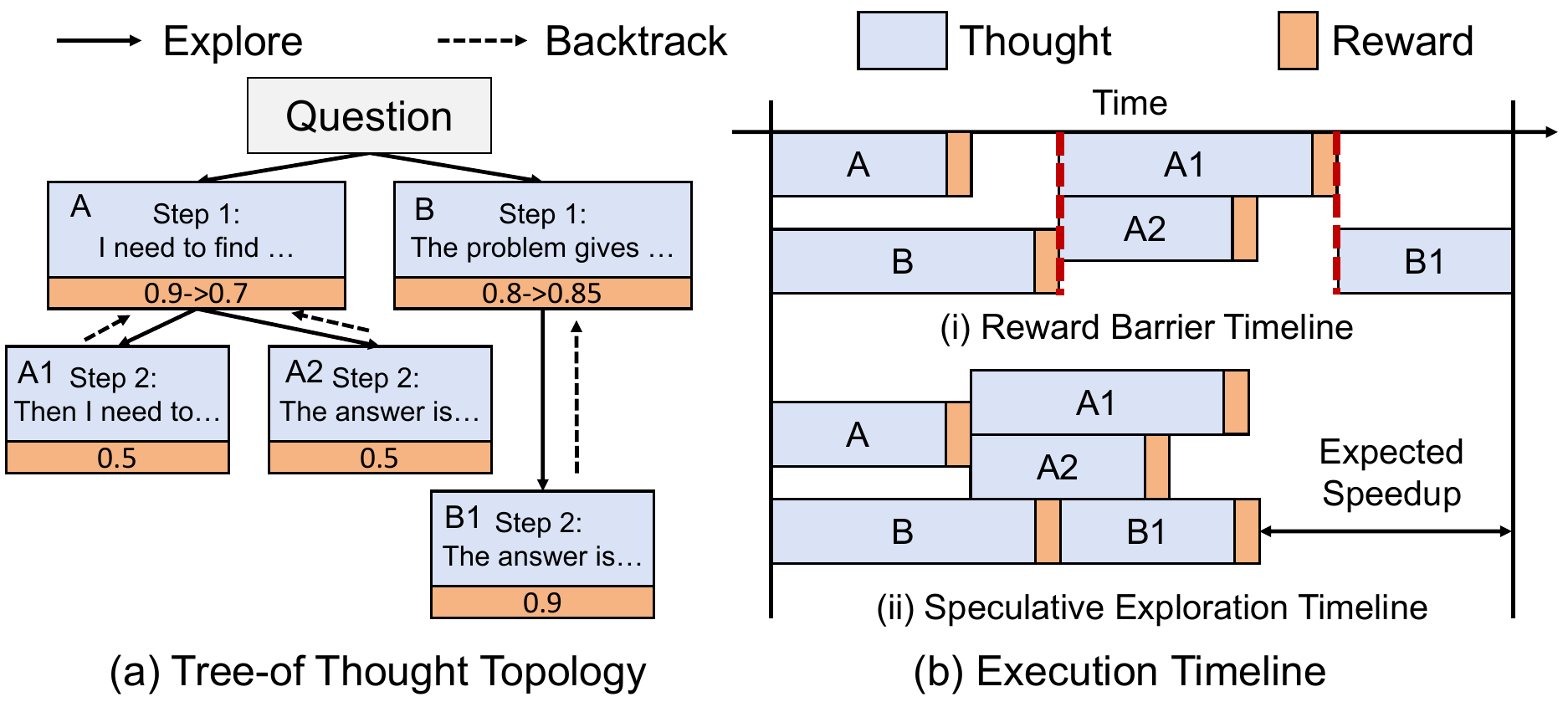}
    \caption{Illustration of ToT and Reward Barrier.}
    \label{fig:intro}
\end{figure}

The recent success of Large Language Models (LLMs) has been driven by their ability to scale with inference-time compute budgets~\cite{snell2024scaling, kojima2022large,qiu2024treebon,team2024qwq,team2025sky,zhangmore,chen2024more,brown2024large}. Two main approaches have been constructed, including Chain-of-Thought (CoT)~\cite{chu2023navigate, wei2022chain,zou2023generalizable,ning2023skeleton} and Tree-of-Thought (ToT)~\cite{beeching2024scaling,cheng2024self,gao2024interpretable, yuan2024advancing,feng2023alphazero,wang2024q}. CoT prompts LLMs to reason over a coherent sequence of tokens (denoted as ``thought'') that serve as intermediate steps toward problem solving. ToT further generalizes CoT by maintaining a search tree and allowing LLMs to explore different branches of the tree adaptively, as in Figure~\ref{fig:intro} (a). To guide exploration, ToT often relies on a reward signal that evaluates existing reasoning branches and decides the next course of action, e.g., expanding existing branches or backtracking. Compared to CoT, ToT enables LLMs to perform complex reasoning and greatly enhance their problem-solving capabilities.

Prior ToT algorithms employ distinct exploration strategies driven by how reward signals guide the search. They can be categorized into two paradigms that impose different dependency patterns on the system \cite{yao2023tree}. \textbf{Breadth-First Search (BFS)} maintains a set of the most promising thoughts at each step, utilizing intermediate reward feedback to concurrently prioritize and expand branches at the current search frontier. In contrast, algorithms like MCTS~\cite{zhang2024rest} incorporate \textbf{Depth-First Search (DFS)}. The reward of all the nodes of a reasoning branch is iteratively updated via reward backpropagation from leaf nodes. The search policy then leverages these refined rewards to revisit and expand prior nodes, enforcing a sequential dependency on previous iterations.

While ToT enhances reasoning capabilities and accuracy, it comes with increased latency overhead \cite{chen2024not,hou2025thinkprune,orches}. Many studies have explored strategies for efficient reasoning. On the algorithmic side, techniques like TrimR focus on reducing the length of thoughts for better reasoning efficiency \cite{lin2025trimr,fu2024efficiently,chen2023mcc, liu2025can,chen2025skip,zhao2025can,wang2025r1,hou2025thinkprune,qi2025optimizing,li2024escape}. On the system side, serving frameworks such as vLLM \cite{kwon2023efficient} and SGLang \cite{zheng2024sglang} achieve significant efficiency improvement with PagedAttention and RadixAttention. However, these system optimizations are primarily designed for linear CoT reasoning and overlook the unique characteristics of ToT reasoning \cite{li2024large}.

In this paper, we identify the critical performance bottleneck in ToT reasoning as the Reward Dependency Barrier as illustrated in Figure~\ref{fig:intro}(b). In Breadth--First Search, the barrier imposes a synchronization overhead: expansion decision is gated by the aggregation of reward values of all concurrent thoughts. High variance in decoding lengths creates stragglers, forcing shorter branches to idle until the longest one completes. For Depth-First Search, the barrier imposes a sequential constraint: the selection of the next trajectory strictly depends on the backpropagated reward from previous iterations, enforcing a serialized execution flow. Consequently, both scenarios prevent the system from sustaining dense batch processing. This inefficiency limits parameter reuse and KV cache sharing, significantly degrading arithmetic intensity and shifting the workload to a memory-bound regime.

To break the barrier, we propose \textbf{\method}~to enhance parallelism in ToT reasoning. By proactively generating subsequent thoughts for promising branches without waiting for the synchronization of reward values, \method{} allows the system to overlap the generation of future thoughts with the completion of current thoughts. It addresses three key challenges in speculative exploration: (i) How to accurately predict which branches to expand speculatively, (ii) How to efficiently allocate computational resources across queries with varying demands and reuse opportunities, and (iii) How to manage skewed tree structures, where a few disproportionately deep thought branches dominate overall inference latency and are difficult to explore speculatively.

To address these challenges, we identify three key insights that shape the design of \method. First, reward stability across iterations enables reliable prediction of high-potential branches, even in DFS. Second, shared computation opportunities, such as KV cache reuse and parameter sharing, can enhance speculative efficiency and resource utilization. Finally, in skewed tree structures, most correct solutions emerge at shallow depths, making it possible to terminate the deep branches early.

Motivated by these insights, SPEX introduces three key techniques:
(i) \textbf{Intra-query speculative branch selection}, which leverages reward stability in DFS and score-based allocation in BFS to identify high-likelihood branches for speculative expansion.
(ii) \textbf{Inter-query budget allocation}, which dynamically distributes speculative resources across queries based on predicted utility, KV cache reuse potential, and system constraints, ensuring balanced and efficient utilization.
(iii) \textbf{Adaptive early termination}, which halts exploration in skewed trees once sufficient confidence is achieved, avoiding unnecessary computation in deep branches.

Finally, to unify speculative exploration across diverse ToT reasoning algorithms, we design SPEX as \textbf{a general producer–consumer execution framework} that integrates the three techniques. The framework decouples branch expansion (producer role) from control logic (consumer role), enabling concurrent progress on both primary and speculative branches while adapting to system resource conditions. 
Our contributions can be summarized as follows:
\begin{itemize}
    \item We identify the reward dependency barrier as the principal bottleneck in ToT reasoning.
    \item We propose SPEX, a speculative exploration framework that breaks this barrier by proactively and speculatively exploring promising branches.
    \item We design three complementary techniques, i.e., intra-query speculative selection, inter-query budget allocation, and early termination strategy, to improve the effectiveness of speculation.
    \item We implement the \textbf{SPEX} on top of SGLang and demonstrate up to $3\times$ speedup over prior-art ToT reasoning algorithms without compromising accuracy. SPEX is available at \href{https://github.com/PKU-SEC-Lab/SPEX}{https://github.com/PKU-SEC-Lab/SPEX}.
\end{itemize}



\section{Background}

\subsection{Tree-of-Thought Reasoning}


Recent research on inference scaling laws has shown that LLM performance improves with increased inference-time compute budget \cite{snell2024scaling, kojima2022large,openai2024learning,openai2025o3mini,R1,team2024qwq,team2025sky,chen2024more,brown2024large}. 
Chain-of-Thought (CoT) extends reasoning depth along a single trajectory, Tree-of-Thought (ToT) further expands reasoning breadth by exploring multiple solution paths in parallel~\cite{yao2023tree}.

Structurally, ToT frames reasoning as a search over a tree of discrete steps. In this framework, each step constitutes a coherent segment of thought, serving as the atomic unit of expansion. Accordingly, a node represents the partial solution state reached by applying a step, and a sequential trajectory of these states from the question to a final answer forms a reasoning path\cite{wu2025inference}.

\begin{figure}[!tb]
        \centering
        \includegraphics[width=0.95\linewidth]{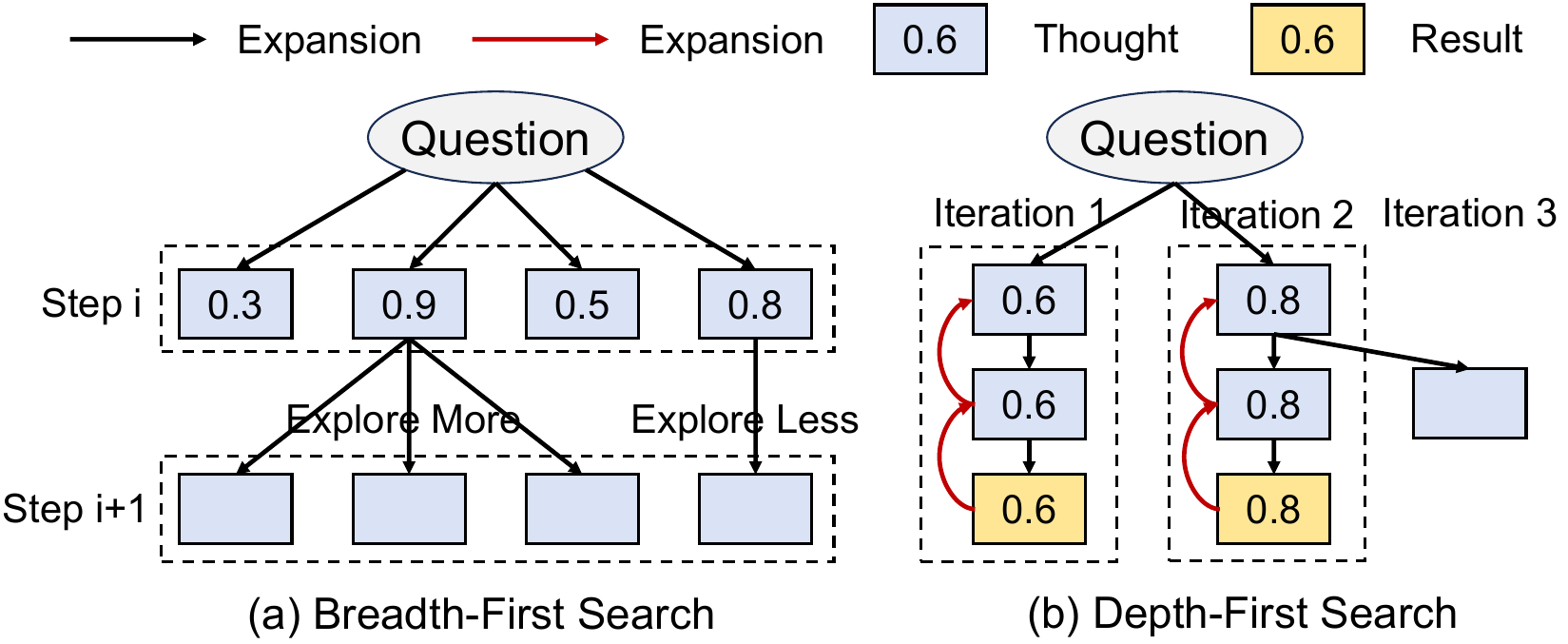}
        \caption{Classification of different ToT algorithms.}
        \label{fig:classification}
    \end{figure}

The search is guided by reward signals, derived either from a dedicated reward model\cite{wu2025inference} or through the reasoning model's intrinsic self-evaluation\cite{fu2025deepthinkconfidence}. As illustrated in Figure~\ref{fig:classification}, ToT algorithms can be categorized by their expansion and traversal strategies into BFS and DFS.


\subsubsection{Breadth-First Search}
Breadth-First Search maintains a collection of the most promising nodes at each step, optimizing the search frontier by utilizing intermediate feedback to concurrently prioritize and extend these branches\cite{yao2023tree}. A representative algorithm is REBASE~\cite{wu2025inference} illustrated in Figure~\ref{fig:classification}(a), which utilizes process rewards to guide this search. It employs a softmax-based mechanism to dynamically allocate expansion budgets.
Formally, given a total sampling budget $B_i$ at depth $i$, the expansion width $W_j$ for a node $n_j$ is calculated as:
\begin{align}
W_j = \text{Round}\left( B_i \frac{\exp(R(n_j)/T_b)}{\sum_k \exp(R(n_k)/T_b)} \right) \nonumber
\end{align}
where $R(N_j)$ denotes the reward score of node $n_j$, $T_b$ is the balance temperature controlling the exploration-exploitation trade-off, and the summation index $k$ iterates over all candidate nodes at the current depth $i$.

\subsubsection{Depth-First Search}
Depth-First Search explores the most promising path until a terminal state is reached, and then leverages updated value estimates to backtrack and revisit high-potential prior states for alternative exploration\cite{yao2023tree}.
Monte Carlo Tree Search (MCTS) is a representative algorithm for this category as illustrated in Figure~\ref{fig:classification}(b)~\cite{chaslot2008monte,xie2024monte,wang2024openr,qi2024mutual}. It balances the trade-off between exploration and exploitation using the Upper Confidence Bound (UCB) metric to select the next action $a^*$:
\begin{align}
a^* = \arg\max_{a} \left[ Q(s, a) + c \cdot \sqrt{\frac{\log N(s)}{N(s, a)}} \right] \nonumber
\end{align}
where $Q(s, a)$ represents the estimated value of taking action $a$ in state $s$, $N(s)$ and $N(s,a)$ denote the visit counts for the state and the specific action, respectively, and $c$ is the exploration constant. By updating these statistics after each traversal, the algorithm retrospectively adjusts its search focus. However, this iterative dependency typically enforces a sequential execution pattern, severely limiting parallelism.

Notably, DFS can operate in isolation or coexist with BFS. Pure DFS, such as RSTAR-MCTS~\cite{qi2024mutual}, generates a single continuous chain per iteration, receiving a reward signal only upon reaching the final outcome. In contrast, REST-MCTS~\cite{zhang2024rest} adopts a hybrid strategy where each iteration functions as a step-wise BFS: at every depth, it expands multiple candidate nodes and utilizes immediate process rewards to select the single best child for the subsequent step.


\subsection{Existing Optimizations for ToT Reasoning}

Existing system optimizations for LLM inference are highly effective for single-path CoT reasoning, but few are specifically designed to support the tree-structured search process of ToT reasoning. This fundamental mismatch leaves current serving infrastructures unable to fully exploit the potential of multi-path search.

On the system side, a series of serving frameworks have substantially improved throughput and latency for linear decoding workloads. 
vLLM \cite{kwon2023efficient} introduces PageAttention, which decouples key-value caching from attention computation.
SGLang \cite{zheng2024sglang} implements RadixAttention to reuse intermediate KV-cache memory across requests. FlashAttention and FlashDecoding utilize the online softmax to reduce memory usage\cite{dao2022flashattention}. FlashInfer integrates the above optimizations of attention into a unified block-sparse framework\cite{ye2025flashinfer}. Together, these efforts demonstrate the effectiveness of system-level optimization for general LLM decoding.

On the algorithmic side, efficient reasoning methods primarily target token efficiency in CoT. Approaches reduce the number of generated tokens or adaptively adjust reasoning length based on task complexity \cite{fu2024efficiently,chen2023mcc, liu2025can,chen2025skip,zhao2025can,wang2025r1,lin2025trimr}. ETS \cite{hooper2025ets} goes beyond token-level analysis by providing a system-oriented perspective, showing that in memory-bound scenarios, KV cache access dominates inference costs. By optimizing KV reuse and access patterns, ETS connects algorithm design with system efficiency.

Despite these advances, existing methods share a fundamental limitation: they are tailored for linear CoT reasoning. While CoT-style optimizations leverage batching and prefix sharing across requests, they do not accommodate the irregular, tree-shaped exploration inherent to ToT reasoning. 

\section{Motivation and Challenge}

Unlike CoT, ToT requires concurrent branch expansion, dynamic allocation of computational budgets, and synchronization based on reward signals. These characteristics introduce new system-level bottlenecks that current inference systems are not designed to handle.

In this section, we first present our key observation that the reward dependency barrier constitutes the primary bottleneck limiting ToT reasoning efficiency (\S\ref{section: motivation}). We then provide a quantitative analysis of how this barrier constrains system performance (\S\ref{sec:motivation-analysis}), followed by our motivation: speculative branch exploration as a means to improve efficiency (\S\ref{sec:motivation-insight}). Finally, we discuss the challenges to perform speculative exploration in ToT reasoning (\S\ref{sec:challenge}).

\subsection{Observation on Reward Dependency Barrier}
\label{section: motivation}

\begin{figure}
    \centering
    \includegraphics[width=0.95\linewidth]{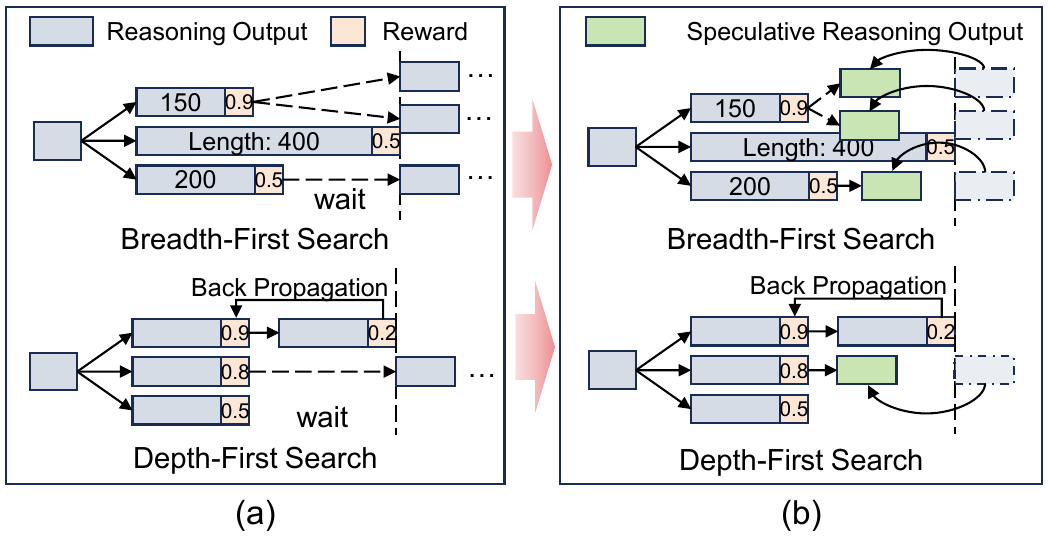}
    \caption{Example of (a) reward barrier and (b) our proposed speculative exploration.}
    \label{fig:key_illustration}
\end{figure}

We observe that the reward dependency barrier is the principal performance bottleneck in ToT reasoning: At each stage of the search, the system must pause to await feedback from the reward model before determining the next action. The manifestation of this barrier varies depending on the underlying algorithmic paradigm and arises from two main sources.

\textbf{1) Intrinsic Sequentiality in Depth-First Search.} 
In DFS algorithm, the search advances along a single reasoning path, with progress contingent on receiving the reward signal. Only after obtaining this feedback can the system decide whether to backtrack and which alternative branch to explore next. This results in inherently sequential execution and severely limits opportunities for parallelism, as depicted in Figure~\ref{fig:key_illustration}(a).

\textbf{2) Synchronization Bottlenecks in Breadth-First Search.} 
In contrast, BFS algorithm can explore multiple candidate branches in parallel at a given search depth. However, its overall progress remains gated by synchronization points at each level: the system must aggregate reward signals across all branches before determining the expansion budget for each node at the next depth, as shown in Figure~\ref{fig:key_illustration}(a). This introduces critical synchronization overheads that limit parallelism, as faster branches are forced to idle until the global aggregation completes.

\begin{figure}[!t]
    \centering
    \includegraphics[width=\linewidth]{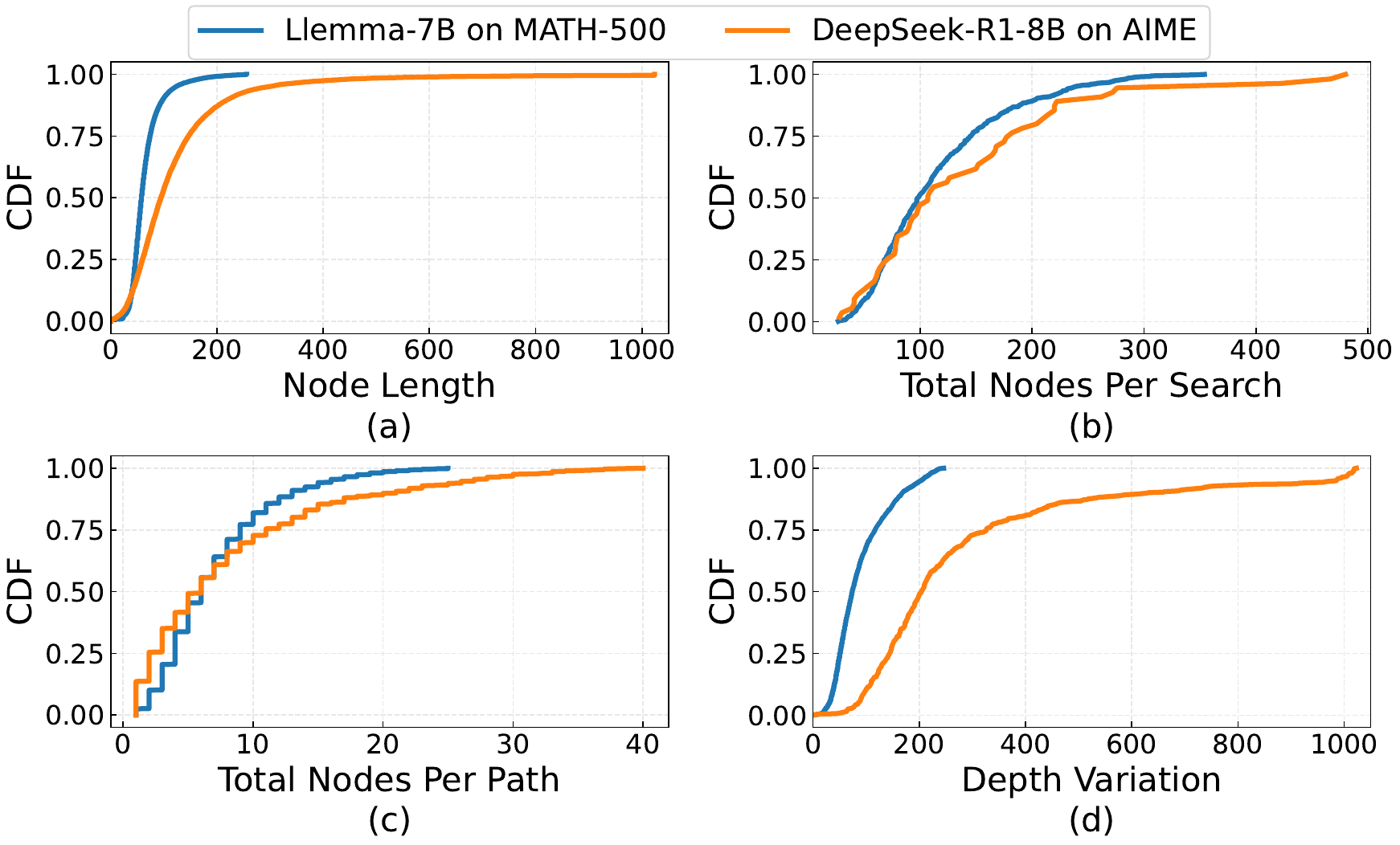}
    \caption{Cumulative Distribution Function (CDF) of reasoning characteristics for Llemma-7B and DeepSeek-R1-8B. (a) Node Lengths; (b) Total Nodes Per Search; (c) Total Nodes Per Path; (d) Variation of Node Lengths at the same depth.}
    \label{fig:trace}
\end{figure}

To quantify the reward barrier, we analyze the cumulative distribution of node lengths and their variation across search depths. We conduct this profiling on two representative workloads: Llemma-7B on MATH-500 and DeepSeek-R1-8B on the challenging AIME benchmark. As shown in Figure~\ref{fig:trace}(a), DFS typically involve a large number of sequential steps, with each step generating 50-100 tokens. Figure~\ref{fig:trace}(b) and~\ref{fig:trace}(c) further illustrate that the total number of nodes per search and the number of nodes along each path are both substantial, indicating that MCTS-style algorithms require traversing many nodes in series. Given that these methods often require hundreds of such steps, this results in significant cumulative latency. For BFS algorithm, Figure~\ref{fig:trace}(d) quantifies the synchronization overhead by plotting the variation in node lengths at the same depth. The slow saturation of the CDF curves reveals a pronounced long-tail effect: while many branches finish early, the system is forced to idle for the slowest "straggler" branches to complete. This high variance, particularly evident in the DeepSeek workload, exacerbates resource underutilization at synchronization barriers.

Overall, this analysis highlights that both algorithmic paradigms are fundamentally constrained by the reward barrier: DFS methods suffer from high sequential latency due to deep search paths, while BFS methods are limited by stragglers and synchronization overhead.

\subsection{System-Level Performance Analysis}
\label{sec:motivation-analysis}

To quantitatively illustrate how the reward dependency barrier limits system efficiency, we analyze the performance of ToT using the Roofline model. Two primary contributors dominate memory operations during inference: (i) model weight accesses and (ii) KV cache accesses. Both are significantly affected when parallelism is constrained, leading to repeated memory operations and inefficient resource utilization.

When batch sizes are small due to limited parallelism, the system must repeatedly load model weights for each reasoning step. Similarly, KV cache accesses become inefficient due to the lack of shared prefixes across reasoning paths. Without sufficient KV sharing, the system repeatedly reloads overlapping or identical context segments across different branches, increasing memory bandwidth consumption. This repeated loading not only adds latency but also prevents tree-based optimizations like RadixAttention.

\begin{figure}[t]
    \centering
    \includegraphics[width=\linewidth]{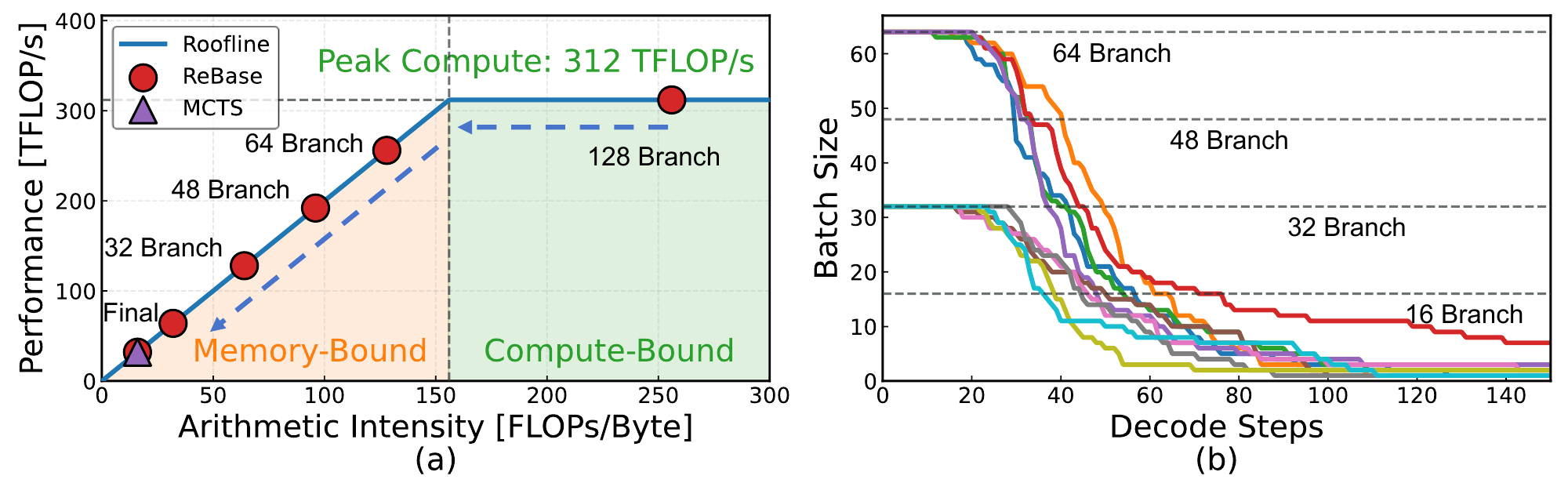}
    \caption{(a) Roofline model analysis; (b) Intensity degradation over decode steps.}
    \label{fig:roofline}
\end{figure}

As shown in Figure~\ref{fig:roofline}(a), these repeated memory operations result in low arithmetic intensity (FLOPs/Byte), making memory access the primary bottleneck.
DFS is inherently confined to the memory-bound region due to their strict sequentiality. While BFS starts with higher parallelism, it suffers from batch attrition. As illustrated in Figure~\ref{fig:roofline}(b), the early completion of shorter branches degrades the effective batch size, dynamically shifting the workload from compute-bound into memory-bound regime.



\subsection{Motivation}
\label{sec:motivation-insight}

While the reward barrier limits inter-branch concurrency, significant latent parallelism exists within the reasoning tree. 
By \textbf{speculatively exploring confident branches} before reward feedback arrives, we can break the reward barrier, reduce synchronization delays, and reuse shared prefixes. 

This motivates \textbf{SPEX}, a speculative exploration framework that exploits potential parallelism to overcome sequential bottlenecks in ToT reasoning.
Specifically, SPEX proactively and speculatively explores promising candidate branches based on predictive signals, as illustrated in Figure~\ref{fig:key_illustration}(b). This removes the blocking dependency on reward feedback, allowing the expansion process to continue without idling.
\subsection{Challenges}
\label{sec:challenge}

However, directly applying speculative exploration is insufficient, as it faces the following challenges.

\textbf{Challenge 1: Speculating Inaccuracy on Future-Needed Branches.}  
Speculating on future-needed branches is straightforward in BFS, where candidate nodes are confined to the current depth. However, for DFS algorithms like MCTS, every node in the tree may be revisited, making it difficult to predict which branches will be required. This complexity increases the difficulty of efficiently allocating speculative computation to accelerate future exploration.

\begin{figure}
    \centering
    \includegraphics[width=\linewidth]{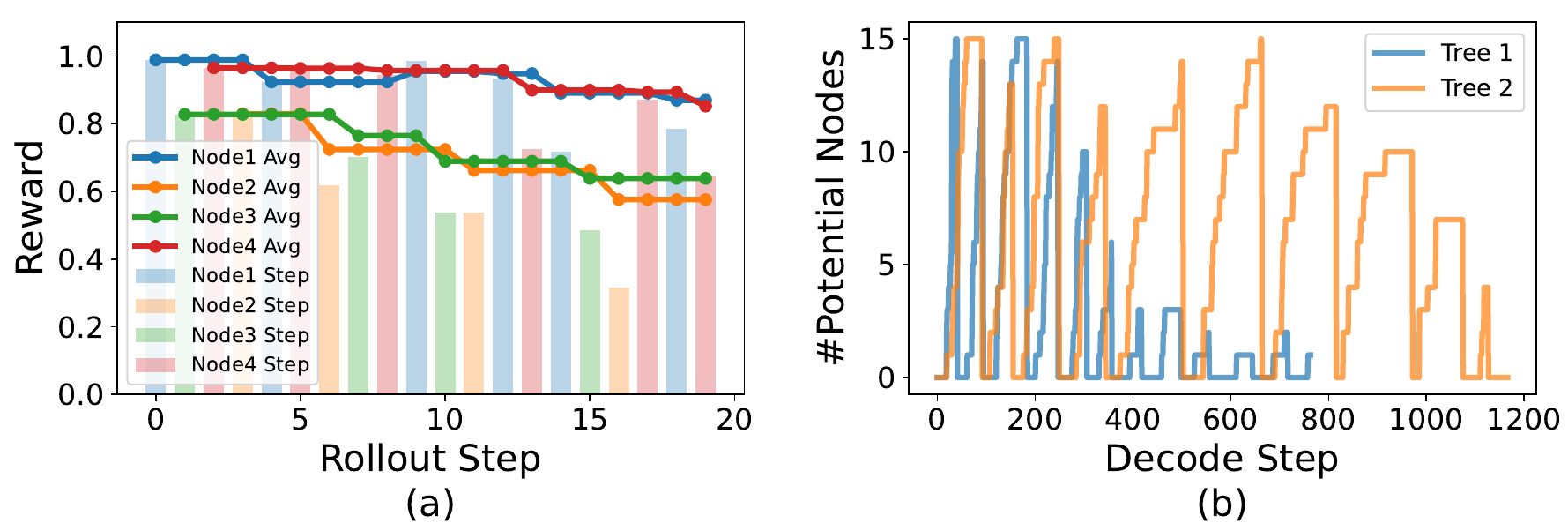}
    \caption{(a) Stability of node rewards in MCTS across rollout steps. (b) Fluctuations in the number of potential speculative nodes for two queries at different decoding steps in REBASE.}
    \label{fig:observation}
\end{figure}

\textbf{Challenge 2:  Complex Speculative Budget Allocation across Multiple Queries.}  
In multi-query scenarios, distributing speculative resources becomes more complex due to variations in the number of potential nodes per query and their probabilities of future relevance. As illustrated in Figure~\ref{fig:observation}(b), the number of nodes available for speculative execution fluctuates significantly between queries. Additionally, these nodes may vary in their likelihood of being utilized, depending on the accuracy of predictions for each query. This variability, combined with limited system resources, requires dynamic allocation strategies that prioritize queries based on their search state, predicted utility, and overall system capacity to optimize throughput and responsiveness.


\begin{figure}
    \centering
    \includegraphics[width=\linewidth]{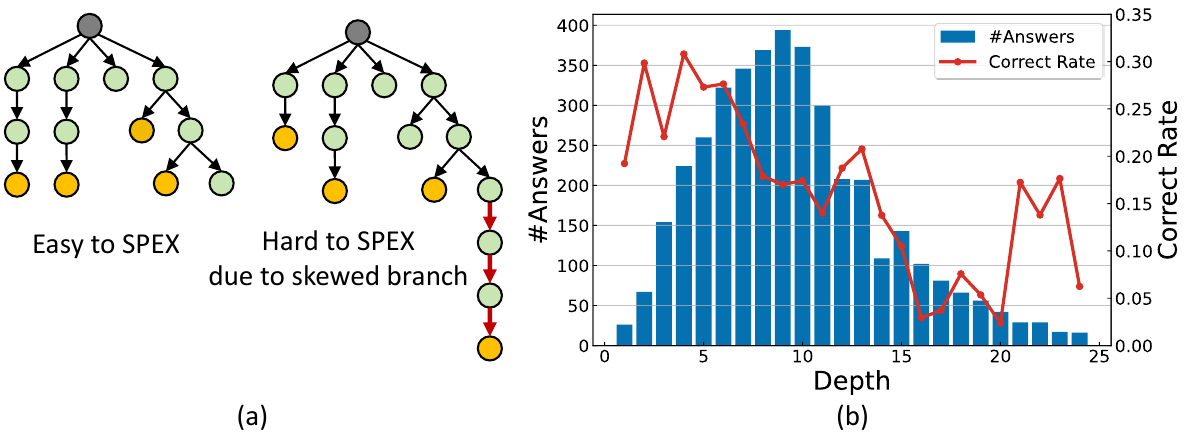}
    \caption{(a) SPEX struggles with skewed trees under the REBASE algorithm due to dominant deep branches. (b) Answer count and correct rate of different reasoning depths in skewed deep trees (depth > 10).}
    \label{fig:skew}
\end{figure}

\textbf{Challenge 3: Inefficient Resource Allocation under Skewed Tree Structures.}  
For highly skewed trees, SPEX struggles to achieve significant acceleration, as shown in Figure~\ref{fig:skew}(a). In such cases, the majority of computational resources are consumed by disproportionately deep branches, which are challenging to predict and explore speculatively. This reduces parallelism, increases latency from sequential exploration, and limits efficient resource allocation, particularly when dealing with long dominant branches.

\section{The SPEX Design}
\label{sec:design}

This section presents the design of SPEX, a speculative exploration framework that breaks the reward dependency barrier by increasing parallelism for ToT reasoning.

\subsection{Overview}
\label{sec:design-overview}

The key idea of SPEX is to speculatively explore confident branches before reward feedback arrives, as presented in \S\ref{sec:motivation-insight}. To address the challenges identified in speculative exploration (\S\ref{sec:challenge}), SPEX leverages three core techniques, each motivated by key insights observed in ToT reasoning. 

\textbf{Insight 1: Reward Stability Enables Prediction.} As shown in Figure~\ref{fig:observation}(a), node rewards in MCTS exhibit high stability across iterations, with reward values changing minimally as rollouts progress. This is because each expansion step updates only one node’s statistics, and rewards, averaged over multiple rollouts, evolve slowly while UCB variations are primarily influenced by visit counts. This stability allows reasonably accurate predictions of node behavior, facilitating speculative execution or pruning strategies to reduce sequential bottlenecks.

Based on the insight, we propose \textit{an efficient intra-query speculative branch selection scheme}, which leverages reward stability in DFS and score-based allocation in BFS to identify high-likelihood branches for speculation (addressing Challenge 1).

\textbf{Insight 2: Shared Computation Shapes Allocation.} Efficient allocation of speculative resources requires evaluating the potential benefits of each query. The primary gains from speculative exploration arise from two forms of shared computation: (i) parameter loading reuse, where speculative branches share model weights with the primary branch, and (ii) KV cache reuse, where branches with overlapping prefixes reuse cached attention states. These benefits are influenced by factors such as the prediction accuracy of speculative branches and the proportion of the KV cache that can be reused. By quantifying these elements, the system can guide resource distribution to maximize the overall utility of speculation while minimizing waste on low-impact queries.

Based on the insight, we propose \textit{an efficient inter-query budget allocation scheme}, which dynamically distributes speculative resources across queries based on predicted utility, KV cache reuse potential, and system constraints, ensuring balanced and efficient utilization (addressing Challenge 2).

\textbf{Insight 3: Answer Depth Bias Favors Shallower Exploration.} As illustrated in Figure~\ref{fig:skew}(b), even for skewed deep trees, most answers are generated at shallower depths, where the correct rate is also higher. This indicates that early-stage exploration typically produces both more frequent and more accurate solutions, suggesting that deep exploration may often be unnecessary under majority vote \cite{li2023making} strategies.

Based on the insight, we propose \textit{an adaptive early termination strategy}, which halts exploration in skewed trees once sufficient confidence is achieved, avoiding unnecessary computation in deep branches (addressing Challenge 3).

\begin{figure}[t]
    \centering
    \includegraphics[width=\linewidth]{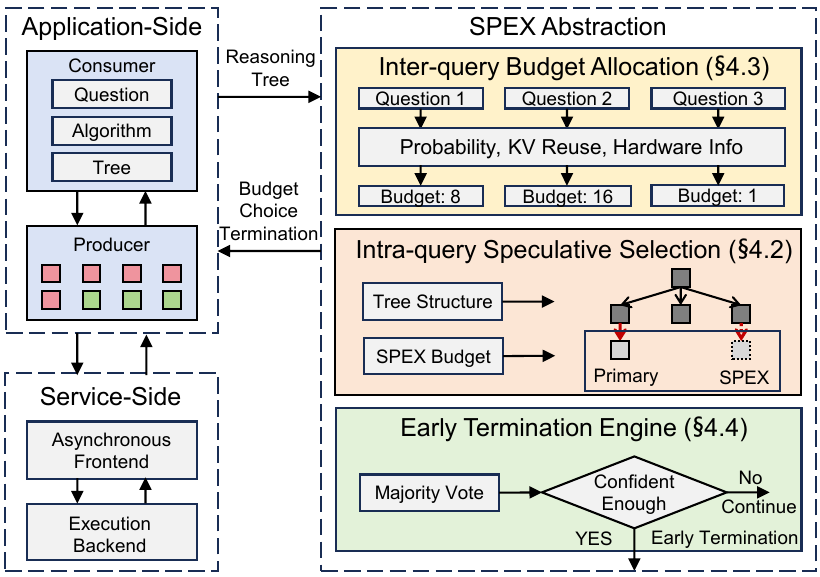}
    \caption{The architectural overview of SPEX.}
    \label{fig:overview}
\end{figure}

Together, these three techniques define the SPEX workflow, as illustrated in Figure~\ref{fig:overview}. First, given multiple queries, SPEX applies \textit{the inter-query budget allocation scheme} to allocate speculative resources based on query probability, KV cache reuse potential, and hardware capabilities (\S\ref{sec:design-inter-query}). Next, for each query, SPEX employs \textit{the intra-query speculative selection scheme} to predict and expand the most promising future branches for speculative exploration (\S\ref{sec:design-intra-query}). Finally, to handle skewed trees, SPEX integrates \textit{the adaptive early termination strategy}, which halts exploration and returns answers when the system has gathered sufficient confidence (\S\ref{sec:design-early-termination}).

Finally, to unify speculative exploration across diverse ToT reasoning algorithms, we design SPEX as \textit{a general producer–consumer execution framework} (\S\ref{sec:design-general-framework}) that integrates the three techniques.

\subsection{Intra-query Speculative Selection}
\label{sec:design-intra-query}

To improve efficiency within a single query, SPEX integrates speculative exploration into the node selection process, employing two complementary strategies: one tailored for DFS and the other for BFS algorithms.

\begin{figure*}
    \centering
    \includegraphics[width=0.97\linewidth]{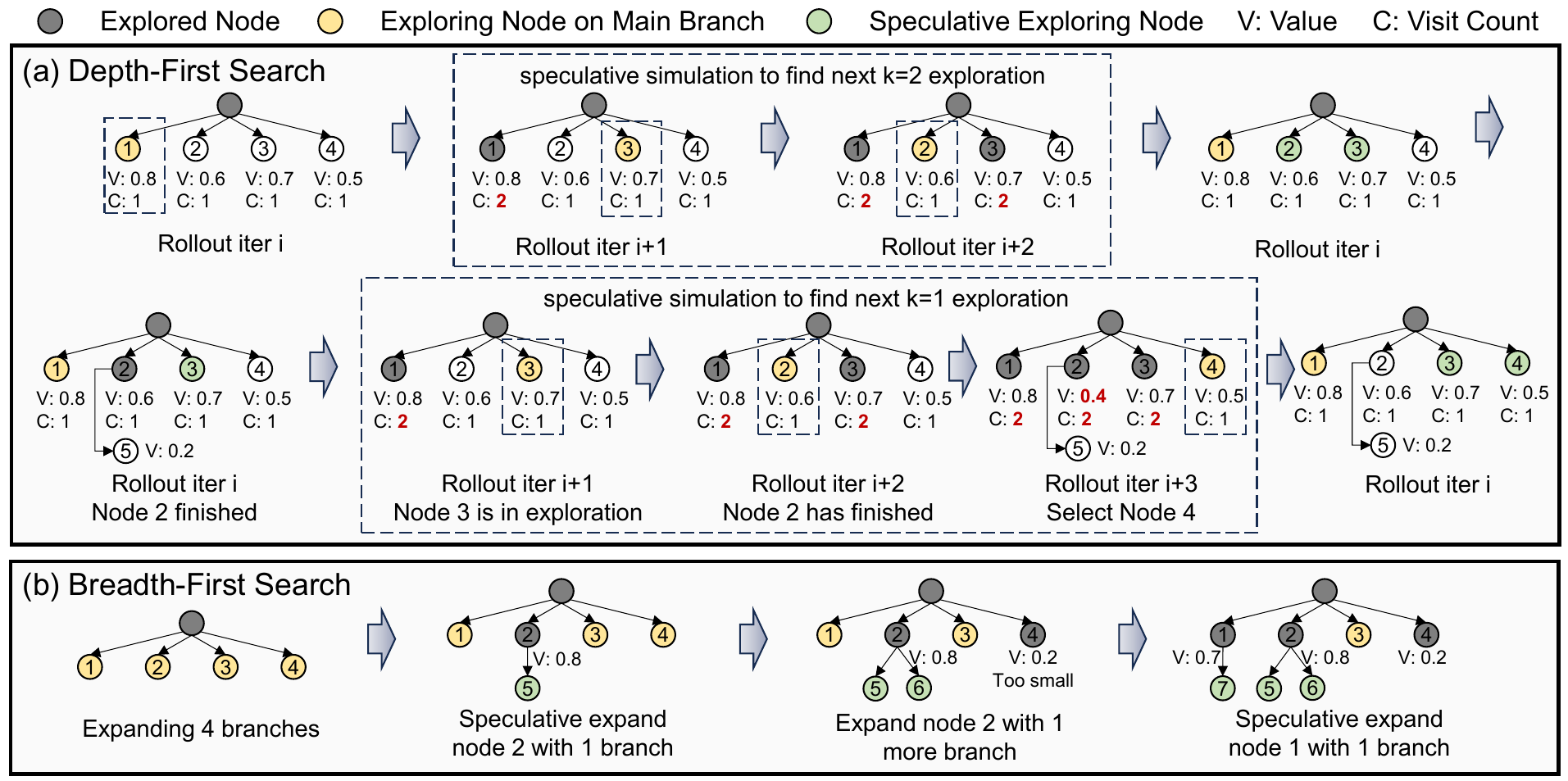}
    \caption{SPEX for (a) DFS and (b) BFS.  
            (a) SPEX simulates the next $k$ iterations, updating visit counts (C) and values (V) after each simulated choice, and skipping nodes already under or completed in speculation.  
            (b) SPEX allocates speculative branches via softmax over reward scores, prioritizing high-value nodes to improve reuse and reduce stragglers.
            }
    \label{fig:illustration}
\end{figure*}

\textbf{1) Speculation for Depth-First Search.}
As shown in Insight~1, reward estimates of individual nodes in MCTS exhibit high stability across iterations, enabling accurate forecasting of future node selections without invoking the reward value.

SPEX leverages this property by \emph{simulating} the next $k$ primary selections under the current tree state.  
After each simulated selection, the algorithm updates visit counts and UCB values before proceeding to the next simulated step, ensuring that subsequent predictions are conditioned on prior choices.  
If the selected node is already under speculative expansion, it is skipped. If speculation for that node has been completed, the stored speculative reward is incorporated into the UCB update to refine the accuracy of future forecasts.  
This process focuses speculative computation on unoccupied, high-likelihood branches while reusing prior results to improve prediction accuracy.

Algorithm~\ref{alg:spec_backtracking} presents the speculative selection procedure, and Figure~\ref{fig:illustration}(a) illustrates its execution. SPEX predicts the next UCB-based choices, skips nodes already under speculation, and integrates completed speculative results into value updates.  
As a result, the system keeps producers busy with reusable work while minimizing redundant expansions.

\begin{algorithm}[t]
    \caption{Speculation for ToT-DFS.}
    \label{alg:spec_backtracking}
    \begin{algorithmic}
        \Require Search tree \textit{tree}, number of speculative branches $k$
        \Ensure Ordered list of speculative nodes to expand 
        \State selected\_nodes $\leftarrow$ [ ]
        \For{$t \gets 1$ to $k$} 
            \State node $\leftarrow$ \textbf{SimulateNext}(\textit{tree})
            \While{node $\in$ active\_expansions \textbf{or} node $\in$ completed\_speculations}
                \If{node $\in$ completed\_speculations}
                    \State \textbf{UpdateUCBWithReward}(node.reward)
                \EndIf
                \State node $\leftarrow$ \textbf{SimulateNext}(\textit{tree})
            \EndWhile
            \State selected\_nodes.append(node)
            \State \textbf{IncrementVisitCount}(node) 
            \State \textbf{UpdateUCB}(\textit{tree})
        \EndFor
        \State \Return selected\_nodes
    \end{algorithmic}
\end{algorithm}

\textbf{2) Speculation for Breadth-First Search.} 
In BFS, speculative exploration targets the inefficiency caused by the long-tail variance in decoding lengths. Since shorter branches finish significantly earlier than the longest ones, they create valuable idle slots at the reward barrier. SPEX leverages this opportunity by proactively expanding these completed short nodes while the system waits for the stragglers.

As illustrated in Figure~\ref{fig:illustration}(b), the speculative budget matches the count of completed nodes. To distribute this budget, SPEX applies the expansion policy of the underlying BFS algorithm to these finished nodes. For instance, with REBASE, SPEX employs its reward-based softmax to determine the expansion width. By strictly adhering to this baseline policy, SPEX allocates more speculative children to higher-scoring nodes (e.g., node 2). This consistency ensures that speculative choices mirror the algorithm's intended search path, thereby maximizing prediction accuracy.




\textbf{Mis-speculation Handling.} SPEX adopts distinct strategies for handling mis-speculation. For DFS, SPEX operates without explicit detection mechanisms. Since DFS explores paths iteratively, a prediction not selected in the current round remains a valid candidate for future rollouts. In contrast, for BFS, SPEX implements strict verification. It detects mis-speculation by cross-referencing predicted nodes with the actual expansion frontier. If a speculated node is not selected by the current frontier, SPEX immediately terminates the corresponding branch to prevent resource wastage.

\subsection{Inter-query Budget Allocation}
\label{sec:design-inter-query}

In multi-query serving, SPEX allocates speculative execution across queries to maximize overall throughput while respecting hardware limits and each query's parallel capacity.

We first determine the global speculative budget $k_{total}$ using a roofline analysis. Achievable throughput is bounded by the intersection of the compute and memory ceilings; the corresponding batch concurrency is taken as $k_{total}$. This ensures that speculation moves execution toward the compute-bound regime without oversaturating on-chip memory bandwidth or degrading arithmetic intensity.

The acceleration of speculation arises from two primary forms of reuse: (i) \emph{parameter reuse}, where speculative branches amortize model parameters with the primary branch, and (ii) \emph{KV cache reuse}, where branches sharing long prefixes with the primary path reuse KV states in attention computation. Let $S_w$ denote the model parameter size, and $S_{KV}(q)$ the amount of KV state in query $q$ that can be reused by its speculative branches.

For each query $q$, we estimate its speculative utility as
\begin{align}
    score_q = C_q \times P_q \times ( S_w + S_{KV}(q)),
\end{align}
where $C_q$ is the exploitable parallel capacity within the query and $P_q$ is the predicted hit rate of speculative branches. 

The allocation of the global budget follows a softmax weighting over these scores:
\begin{align}
    k_q = \min \bigg( C_q,\; \big\lfloor k_{total} \cdot 
    \frac{e^{\tau\, score_q}}{\sum_{q'} e^{\tau\, score_{q'}}} \big\rfloor \bigg),
\end{align}
where $\tau$ controls allocation sharpness. This strategy assigns more speculative capacity to queries with higher expected benefit while avoiding overspeculation beyond each query’s intrinsic parallelism.

\subsection{Early Termination for Skewed Trees}  
\label{sec:design-early-termination}

In highly skewed trees, SPEX faces significant challenges in achieving acceleration due to the disproportionate computational cost of exploring deep branches. However, as noted in Insight 3, most answers are generated at shallower depths, where the correctness rate is higher. This suggests that exhaustive exploration of deep branches is often unnecessary, especially when reliable solutions can already be inferred from earlier exploration stages.  

To address this, we propose an \textit{adaptive early termination strategy} that halts exploration once sufficient evidence has been gathered to confidently determine a result. Specifically, during exploration, SPEX monitors the total number of generated answers $n$ and the confidence scores of the top two hypotheses. The confidence scores are defined as:
\begin{align}
\text{conf}_{\text{1st}} = {\sum_{i \in \text{1st}} w_i}, \quad \text{conf}_{\text{2nd}} = {\sum_{i \in \text{2nd}} w_i},
\end{align}
where $\text{conf}_{\text{1st}}$ and $\text{conf}_{\text{2nd}}$ represent the total rewards belonging to the top 1 and top 2 answers, respectively.

Exploration terminates early if it satisfies the condition:
\begin{align}
n \geq t \quad \text{and} \quad \text{conf}_{\text{1st}} - \text{conf}_{\text{2nd}} > \alpha \cdot \text{avg}(\text{w}_{\text{2nd}}),
\end{align}
where $t$ is the minimum number of answers required, $\alpha$ is a scaling factor controlling the margin of confidence, and $\text{avg}(\text{w}_{\text{2nd}})$ is the average weight of answers belonging to the second-best hypothesis. In the experiments, $\alpha$ is set to 0.5.
This ensures that exploration avoids unnecessary computation in deep branches when high-confidence solutions can already be determined from shallower exploration. 

\begin{algorithm}[t]
    \caption{SPEX Producer-Consumer Framework}
    \label{alg:framework}
    \begin{algorithmic}
        \Require Initial state $s_0$, search algorithm type $\mathcal{A}$
        \Ensure Final state $s_n$
        \State tree $\leftarrow$ Tree structure with root $s_0$
        \State request\_queue, completion\_queue $\leftarrow$ []
        \State Launch multiple ProducerWorker() instances
        \State request\_queue.put($s_{0}$)
        \While{search is not complete}
            \State $(node, child) \leftarrow$ \textbf{await} completion\_queue
            \If{tree.is\_on\_primary\_branch ($node, \mathcal{A}$)}
                \State tree.add\_node(child)
                \State tree.expand($\mathcal{A}$)
            \Else
                \State tree.add\_spec\_node(child)
            \EndIf
            \While{has\_idle\_producers()}
                \State spec\_node $\leftarrow$ tree.select\_spec()
                \State request\_queue.put(spec\_node)
            \EndWhile
        \EndWhile
        \State Return tree.$s_n$
        \State
        \Procedure{ProducerWorker}{}
            \While{not Terminate}
                \State node $\leftarrow$ \textbf{await} request\_queue
                \State new\_child $\leftarrow$ expand\_node(node)
                \State completion\_queue.put(node, new\_child)
            \EndWhile
        \EndProcedure

    \end{algorithmic}
\end{algorithm}

\subsection{General Producer–Consumer Framework}
\label{sec:design-general-framework}

To unify speculative exploration across different ToT algorithms, we design SPEX as \textit{a general producer–consumer execution framework}. It decouples branch expansion (producer role) from control logic (consumer role), enabling concurrent processing of both primary and speculative branches while dynamically adapting to system conditions.

Specifically, multiple producers execute explorations in parallel. A producer may be assigned to (i) a node along the \textit{primary branch}—the branch that the underlying search algorithm $\mathcal{A}$ is currently committed to exploring—or (ii) a node from \textit{speculative branches} predicted to have high utility. The centralized consumer loop monitors the completion of these expansions through a shared completion queue.

When a completed node belongs to the primary branch, the consumer immediately appends it to the search tree and triggers the next expansion step dictated by $\mathcal{A}$. Simultaneously, the consumer evaluates whether idle producers can be utilized for additional speculative expansions, thereby overlapping useful computation with reward model latency.
If the completed node originates from a speculative branch, it is appended to the speculative subtree. The system again inspects available resources and, if capacity permits, issues further speculative expansions to keep producer threads saturated.

Algorithm~\ref{alg:framework} outlines the execution flow. 
The design achieves three benefits.
\textbf{1) Non-blocking Execution:} Primary branch progress is never stalled by speculative work.
\textbf{2) Opportunistic Parallelism:} Idle resources are opportunistically exploited for parallel speculative exploration.
\textbf{3) Algorithm-agnostic Unification:} The framework remains agnostic to the underlying reasoning algorithm, supporting different paradigms within a unified scheduling architecture.

\section{Implementation}

We implemented the \textbf{SPEX system} on top of SGLang \cite{zheng2024sglang}, introducing several key modifications and components.

On the server side, we modified the SGLang front-end implementation. The original system was designed to handle batch requests that return results collectively. To enable finer-grained scheduling, we re-engineered the system to support asynchronous responses, allowing individual requests to return results independently.

On the application side, we utilized Python's coroutine framework to implement the coordinated operation of a \textbf{Producer-Consumer} model, which manages interactions with SGLang. All three core components of the SPEX system were implemented on the application side, ensuring seamless integration and efficient task execution.
\begin{figure*}
    \centering
    \includegraphics[width=\linewidth]{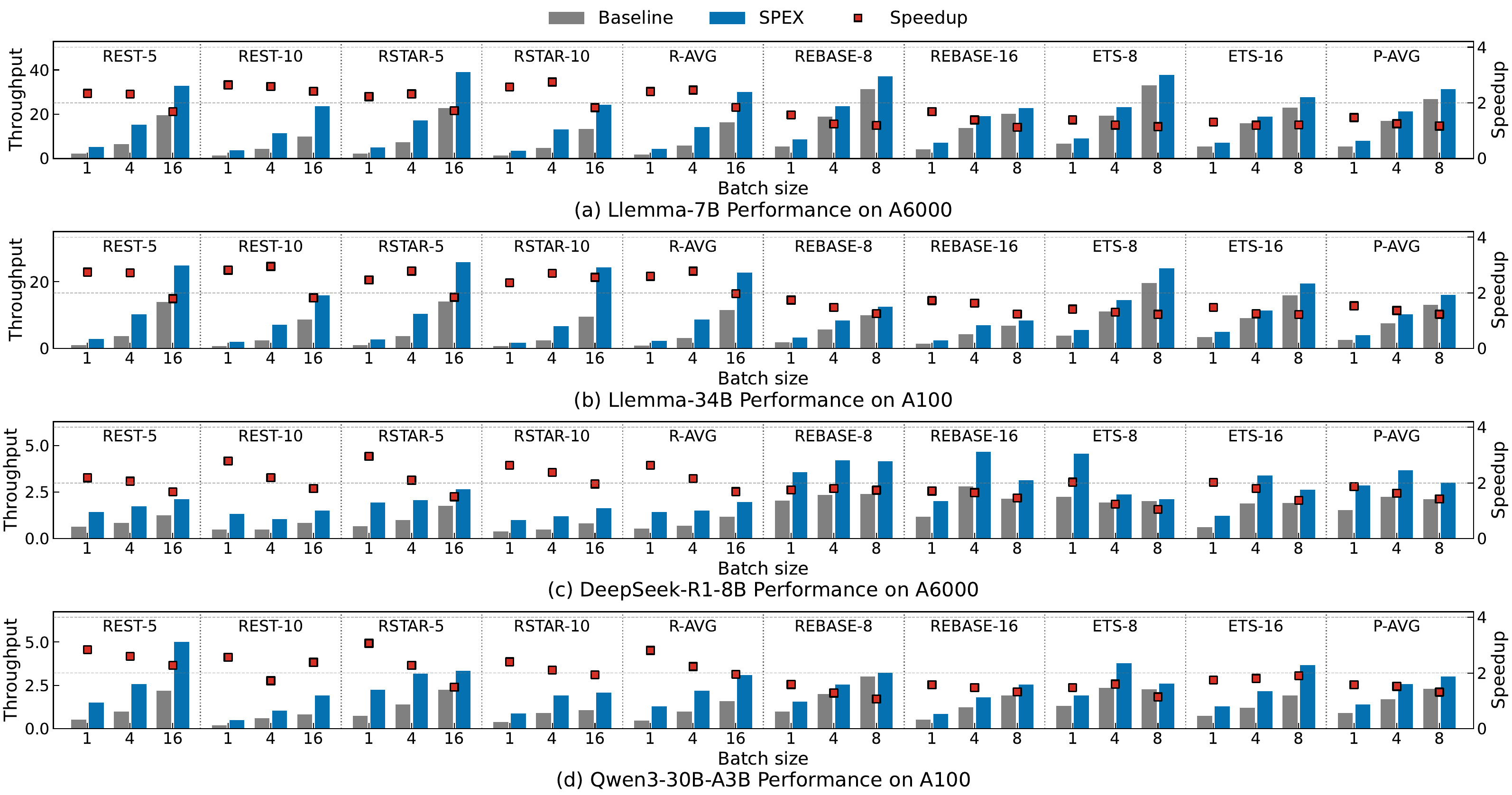}
    \caption{The speedup and throughput (finished questions per minute) for different ToT reasoning tasks.}
    \label{fig:main_perf}
\end{figure*}

\section{Experimental Evaluation}


\subsection{Experimental Setup}

\textbf{Models.} We evaluated SPEX across four models: Llemma-7B, Llemma-34B~\cite{azerbayev2023llemma}, DeepSeek-R1-8B\cite{R1}, and Qwen3-30B-A3B\cite{yang2025qwen3technicalreport}. The Llemma suite was selected to ensure comparability with established ToT baselines, as these models are used in those evaluations\cite{wu2025inference,hooper2025ets}. We employ its dedicated reward model for evaluation.
Conversely, DeepSeek and Qwen were included to validate generalization to state-of-the-art models, where we derived reward signals directly from intrinsic log probabilities to serve as confidence proxies\cite{fu2025deepthinkconfidence}.

\textbf{Hardware.} We conducted experiments across different hardware setups to evaluate SPEX’s performance under various scenarios. Specifically, we tested Llemma-7B and Deepseek-8B on NVIDIA A6000 GPUs, while Llemma-34B and Qwen3-30B-A3B were evaluated on NVIDIA A100 GPUs. These setups allowed us to analyze and compare performance across diverse compute environments.

\textbf{Datasets.} To ensure a fair comparison, we matched the dataset difficulty to the capabilities of each model. For the Llemma family, we utilized GSM8K~\cite{cobbe2021training} and MATH-500~\cite{hendrycks2021measuring,lightman2023let} following the experiment setup of REBASE and ETS, as more extreme challenges like AIME remain intractable for Llemma even with ToT. In contrast, for the state-of-the-art models (DeepSeek and Qwen), simpler tasks like MATH-500 can be easily solved without search. Therefore, we utilized high-difficulty benchmarks, including AIME 2024\cite{AIME2024}, AIME 2025\cite{AIME2025}, BRUMO\cite{BRUMO2025}, and HMMT\cite{HMMT2025} to demonstrate the performance of our framework.

\begin{table}
    \centering
    \small
    \caption{Configurations of evaluated algorithms. The configuration values denote the target number of generated answers.}
    \begin{tabular}{ccc}\toprule
         Algorithm &Type& Configuration\\\midrule
         REST-MCTS\cite{zhang2024rest} & DFS+BFS & 5, 10\\
         RSTAR-MCTS\cite{qi2024mutual} & DFS & 5, 10\\
         REBASE\cite{wu2025inference} & BFS & 8, 16\\
         ETS\cite{hooper2025ets} & BFS & 8, 16\\ \bottomrule
    \end{tabular}
    \label{tab:exp_alos}
\end{table}

\textbf{Algorithms.} As listed in Table~\ref{tab:exp_alos}, we evaluated SPEX on four representative ToT algorithms: REST-MCTS, RSTAR-MCTS, REBASE and ETS. Each algorithm was tested under two configurations, which represent the number of answers to generate; specifically, this corresponds to the rollout count for DFS methods and the initial search width for BFS methods.

\subsection{End-to-End Performance}

Figure~\ref{fig:main_perf} presents the throughput and speedup of SPEX compared to baseline across various algorithms, configurations, and batch sizes. The results demonstrate consistent performance improvements with SPEX under diverse settings.

On average, SPEX achieves a speedup of 1.8–3$\times$ for DFS and 1.2–1.9$\times$ for BFS. The greater acceleration for DFS stems from their inherently limited branch-level parallelism, where SPEX introduces more opportunities for optimization. In contrast, algorithms without DFS like REBASE already exhibit significant parallelism, leaving less room for further improvement. Among DFS algorithms, REST-MCTS achieves slightly higher speedup compared to RSTAR-MCTS, as the BFS creates more opportunities for SPEX optimization.

\textbf{Impact of Batch Size}: The speedup is particularly notable at smaller batch sizes. For DFS algorithms, SPEX can achieve up to 3× acceleration at small batch sizes, and for BFS-only algorithms, the speedup reaches up to 1.7$\times$. As the batch size increases, the system transitions from being memory-bound to compute-bound, reducing SPEX's relative benefits.

\textbf{Impact of Configuration}: DFS algorithms demonstrate higher speedup when generating more answers, as this increases the number of potential nodes. Conversely, for BFS-only algorithms, generating more answers results in lower speedup. This is because algorithms like REBASE expand branches proportional to the required answers, transitioning to a compute-bound regime.

\begin{table}[t]
\caption{Accuracy evaluation. }
\label{table:acc}
\centering
\scriptsize
\resizebox{\linewidth}{!}{
\begin{tblr}{
  colspec={Q[c,wd=7mm] Q[c,wd=7mm] c c c c c c c},
  row{1} = {font=\bfseries},
  rowhead=2,
  hlines,
}
Model       & Dataset  & Method   & REST        & RSTAR             & REBASE             & ETS                & AVG \\
Config.     &          &          & 5 / 10      & 5 / 10            & 8 / 16             & 8 / 16             &     \\
\SetCell[r=4]{c}Llemma\newline -7B
            & \SetCell[r=2]{c}MATH\newline-500 & Baseline &
35.2 / 37.4 & \textbf{35.6 / 37.0} & \textbf{43.4} / 45.2 & 40.6 / 43.8       & 40.3 \\
            &                         & SPEX     &
\textbf{36.8 / 38.8} & 33.8 / 36.4          & 40.8 / \textbf{46.4} & \textbf{42.2 / 46.0} & \textbf{41.5} \\

            & \SetCell[r=2]{c}GSM8K   & Baseline &
75.8 / 77.1 & 74.6 / 76.4           & 83.5 / 86.7          & 82.1 / 83.4          & 80.7 \\
            &                         & SPEX     &
\textbf{77.3 / 78.1} & \textbf{74.9 / 76.8} & \textbf{84.6 / 87.0} & \textbf{85.0 / 86.7} & \textbf{81.8} \\

\SetCell[r=4]{c}Llemma\newline -34B
            & \SetCell[r=2]{c}MATH\newline-500 & Baseline &
\textbf{41.2 / 40.0} & \textbf{37.2} / 38.4  & 43.4 / \textbf{48.0} & 41.0 / \textbf{45.6} & 42.3 \\
            &                         & SPEX     &
38.2 / \textbf{40.8} & 34.8 / \textbf{38.8} & \textbf{46.0} / 46.6  & \textbf{44.9 / 45.4} & \textbf{42.9} \\

            & \SetCell[r=2]{c}GSM8K   & Baseline &
79.9 / 80.9 & 77.7 / 80.1           & 87.3 / 88.0          & \textbf{86.7 / 87.3}  & 83.9 \\
            &                         & SPEX     &
\textbf{80.0 / 81.7} & \textbf{79.1 / 80.3} & \textbf{87.6 / 88.0} & 86.5 / \textbf{87.9} & \textbf{84.2} \\

\SetCell[r=8]{c}Deepseek
            & \SetCell[r=2]{c}AIME24  & Baseline &
73.3 / \textbf{73.3} & 66.7 / 70.0           & \textbf{80.0} / 76.7 & 73.3 / 73.3          & 73.3 \\
            &                         & SPEX     &
\textbf{76.7 / 70.0} & \textbf{66.7 / 73.3} & 76.7 / \textbf{76.7} & \textbf{76.7 / 73.3} & \textbf{73.8} \\

            & \SetCell[r=2]{c}AIME25  & Baseline &
70.0 / \textbf{73.3} & 70.0 / 70.0           & 66.7 / 63.3          & 60.0 / 60.0          & 66.7 \\
            &                         & SPEX     &
\textbf{70.0 / 70.0} & \textbf{70.0 / 73.3} & \textbf{66.7 / 70.0} & \textbf{63.3 / 66.7} & \textbf{68.8} \\

            & \SetCell[r=2]{c}BRUMO & Baseline &
70.0 / \textbf{70.0} & 73.3 / 73.3           & 70.0 / 73.3          & 63.3 / 63.3          & 69.6 \\
            &                         & SPEX     &
\textbf{70.0 / 66.7} & \textbf{76.7 / 76.7} & \textbf{76.7 / 73.3} & \textbf{70.0 / 70.0} & \textbf{72.5} \\

            & \SetCell[r=2]{c}HMMT  & Baseline &
60.0 / 60.0 & 60.0 / 60.0           & \textbf{60.0 / 60.0} & 53.3 / 60.0          & 59.2 \\
            &                         & SPEX     &
\textbf{60.0 / 60.0} & \textbf{60.0 / 63.3} & 56.7 / 56.7          & \textbf{60.0 / 60.0} & \textbf{59.6} \\

\SetCell[r=8]{c}Qwen
            & \SetCell[r=2]{l}AIME24  & Baseline &
70.0 / 73.3 & 70.0 / 73.3           & \textbf{80.0 / 76.7} & 73.3 / 73.3          & 73.8 \\
            &                         & SPEX     &
\textbf{73.3 / 80.0} & \textbf{80.0 / 80.0} & 76.7 / \textbf{76.7} & \textbf{76.7 / 80.0} & \textbf{77.9} \\

            & \SetCell[r=2]{l}AIME25  & Baseline &
\textbf{70.0 / 66.7} & 66.7 / 66.7           & 60.0 / 63.3          & 63.3 / \textbf{60.0} & 64.6 \\
            &                         & SPEX     &
63.3 / \textbf{70.0} & \textbf{66.7 / 66.7} & \textbf{70.0 / 70.0} & \textbf{63.3 / 56.7} & \textbf{65.8} \\

            & \SetCell[r=2]{l}BRUMO   & Baseline &
\textbf{73.3 / 76.7} & 76.7 / 73.3           & \textbf{73.3 / 76.7} & 70.0 / 73.3          & 74.2 \\
            &                         & SPEX     &
70.0 / \textbf{76.7} & \textbf{80.0 / 73.3} & 70.0 / \textbf{76.7} & \textbf{76.7 / 76.7} & \textbf{74.6} \\

            & \SetCell[r=2]{l}HMMT    & Baseline &
60.0 / 60.0 & 56.7 / 60.0           & 63.3 / 56.7          & 56.7 / \textbf{63.3} & 59.6 \\
            &                         & SPEX     &
\textbf{63.3 / 60.0} & \textbf{60.0 / 60.0} & \textbf{63.3 / 60.0} & \textbf{60.0 / 60.0} & \textbf{60.8} \\
\end{tblr}
}
\end{table}

\subsection{Accuracy Evaluation}
Table~\ref{table:acc} demonstrates the accuracy evaluation of SPEX compared to the baseline across different algorithms and datasets. The results indicate that early termination introduced by SPEX does not negatively impact accuracy. For most configurations, SPEX can achieve slightly higher accuracy than the baseline. This gain stems from our adaptive early termination technique. Deep, skewed branches tend to exhibit lower correctness. By terminating these branches early upon reaching high confidence, SPEX prevents low-quality tails from skewing the majority vote.

\subsection{Ablation Study}
\begin{figure}
    \centering
    \includegraphics[width=\linewidth]{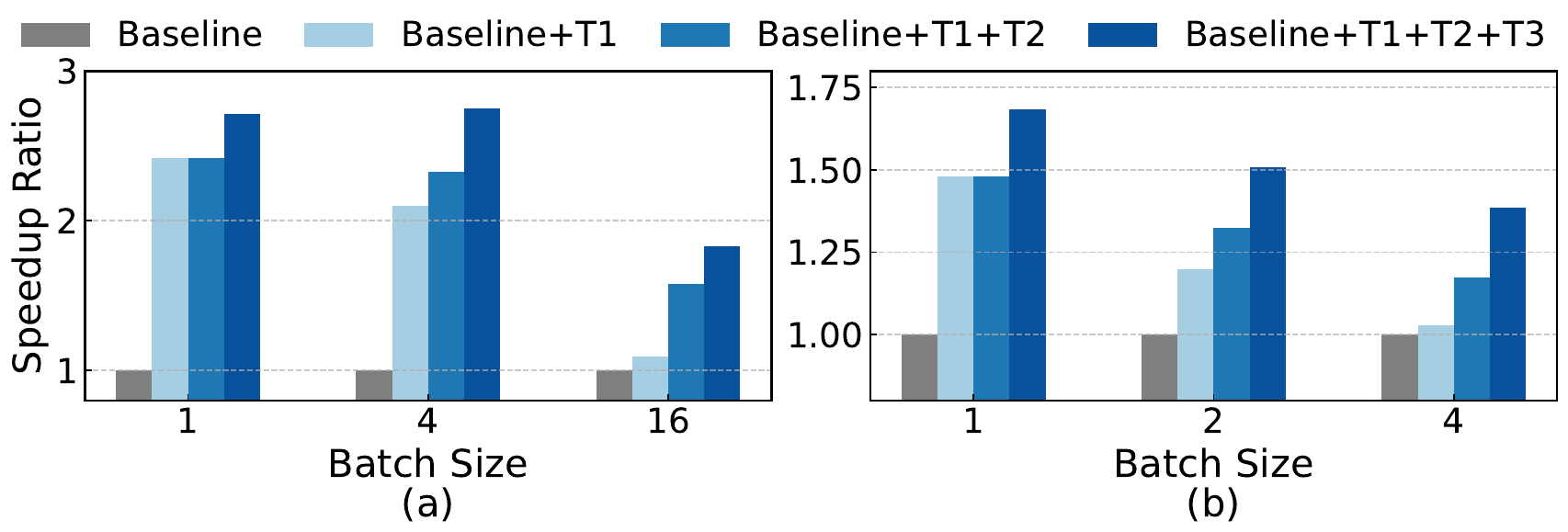}
    \caption{Ablation study of three techniques in SPEX for (a) RSTAR-10 and (b) REBASE-16. T1 represents Intra-query Speculative Selection, T2 represents Inter-query Budget Allocation and T3 represents Early Termination. }
    \label{fig:ablation}
\end{figure}

To evaluate the contribution of each key technique in SPEX, we conducted an ablation study as shown in Figure~\ref{fig:ablation}. Three techniques were analyzed: T1 (Intra-query Speculative Selection), T2 (Inter-query Budget Allocation), and T3 (Early Termination).

T1 demonstrates significant speedup by allowing speculative execution within queries to exploit parallelism, especially for smaller batch sizes. T2 becomes increasingly critical as batch size grows, ensuring efficient allocation of computational resources when the system transitions from memory-bound to compute-bound. T3 provides a stable speedup of around 1.2$\times$ across all configurations by identifying and terminating low-potential branches early.

The combination of all three techniques yields the highest overall performance, with notable improvements in both small-batch and large-batch scenarios.

\subsection{Compatibility with Speculative Decoding}

\begin{figure}
    \centering
    \includegraphics[width=\linewidth]{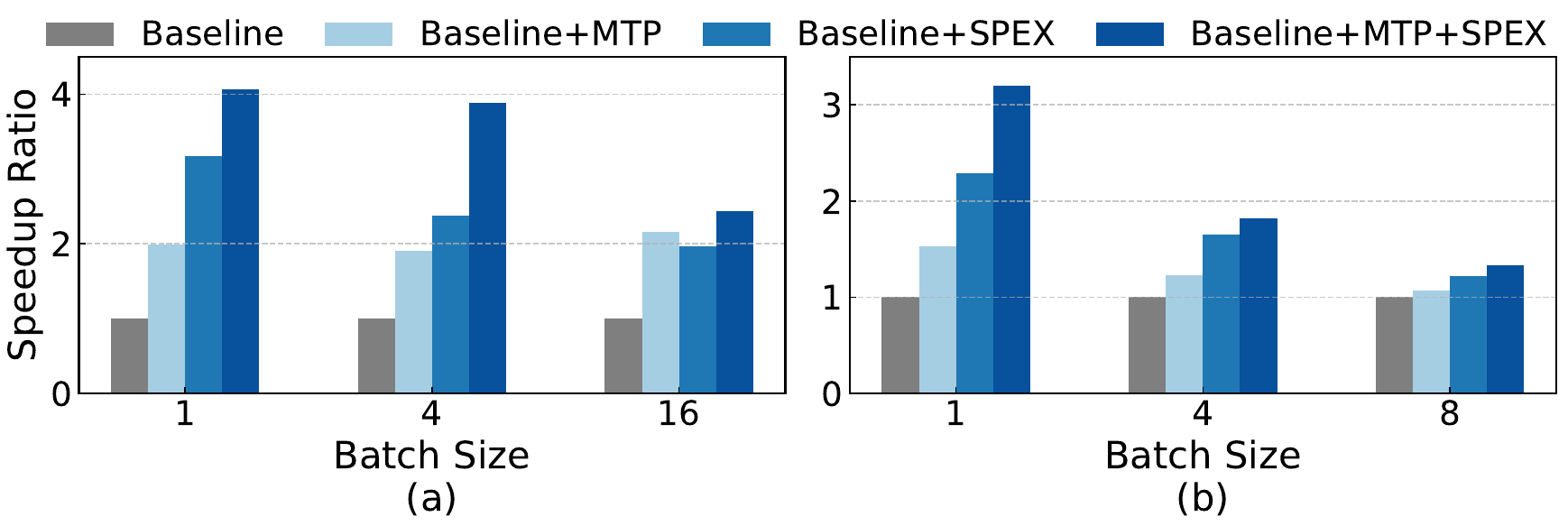}
    \caption{Orthogonality analysis of \method{} and MTP for (a) RSTAR-10 and (b) REBASE-8.}
    \label{fig:mtp}
\end{figure}

To demonstrate that \method{} is orthogonal to token-level optimizations, we integrated it with the Multi-Token Prediction (MTP) module of DeepSeek-R1-8B.
Since the performance of speculative decoding is highly sensitive to the token tree, we empirically tuned the token tree size based on the batch size. For RSTAR-10, which inherently has a smaller effective batch size due to sequential dependency, we maintained a token tree size of 16 across all settings. For REBASE-8, we adopted an adaptive configuration: a tree size of 16 for batch size (BS) 1, scaled down to 8 for BS=4, and 4 for BS=8, to mitigate compute contention.

Figure~\ref{fig:mtp} illustrates the speedup breakdown. The results confirm that \method{} and MTP target distinct bottlenecks and can be effectively composed. The combination (\textbf{Baseline+MTP+SPEX}) consistently yields the highest throughput. Notably, for RSTAR-10 at BS=1, combining SPEX with MTP boosts the speedup from $\sim2.0\times$ (MTP only) and $\sim3.1\times$ (SPEX only) to a remarkably $\sim4.1\times$. As the batch size increases, the system transitions from memory-bound to compute-bound. Consequently, the marginal gain from MTP diminishes. However, \method{} maintains robust acceleration even when token-level speculation saturates. 

\subsection{Overhead analysis}

\begin{figure}
    \centering
    \includegraphics[width=\linewidth]{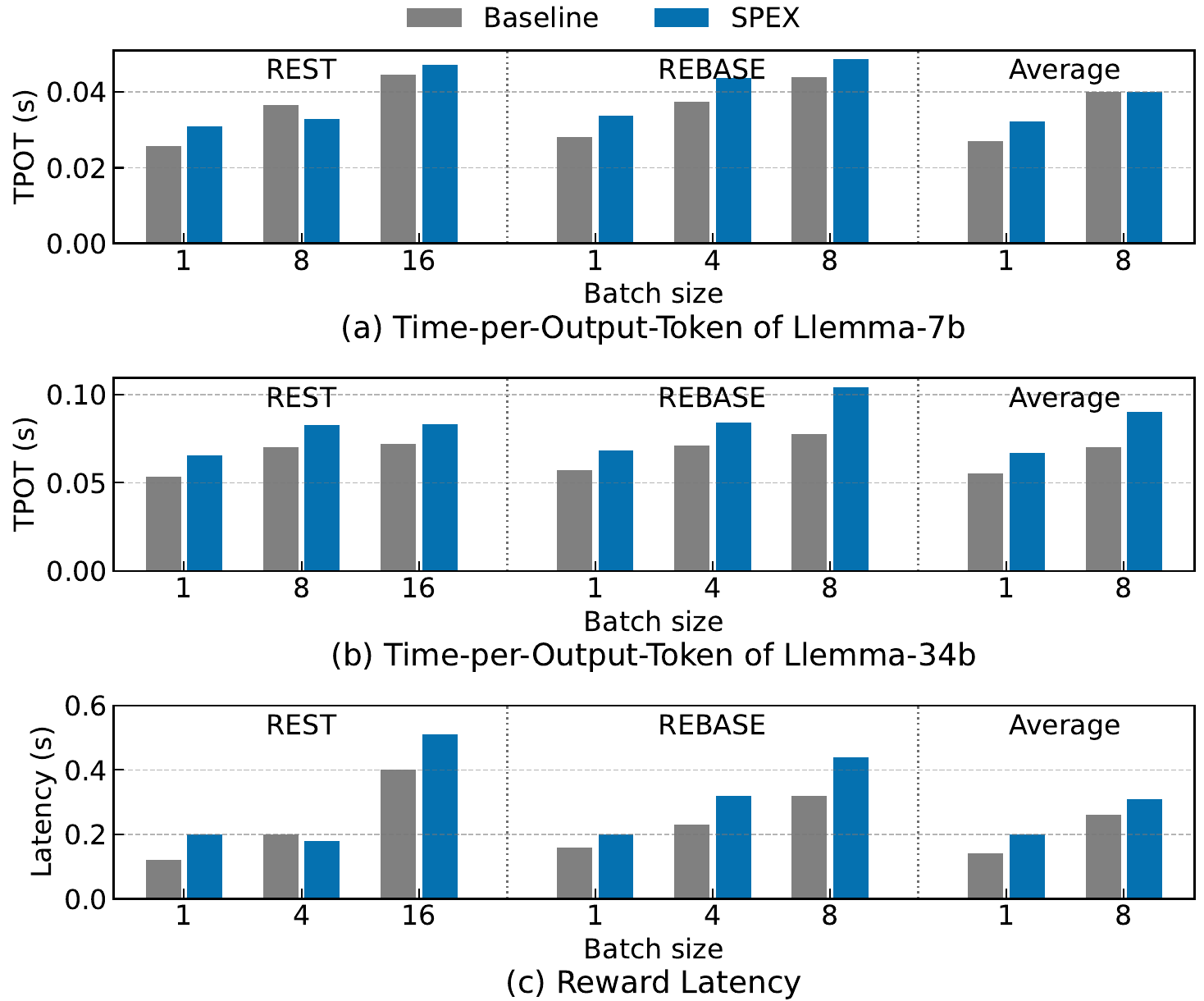}
    \caption{Overhead analysis of SPEX.}
    \label{fig:overhead}
\end{figure}

As SPEX increases the number of parallel branches within each query, it introduces additional overhead that affects both the reasoning model's Time-per-Output-Token (TPOT) and the reward model's evaluation latency, particularly when the system approaches a compute-bound regime. Figure~\ref{fig:overhead} illustrates these effects.

For the reasoning model's output speed, while SPEX raises the risk of pushing the system into an over-compute-bound state, the observed overhead remains controlled within 15\%. This indicates that SPEX effectively balances the trade-off between batch size scaling and computational efficiency.

For the reward model's evaluation latency, the impact of SPEX is minimal. The average delay introduced by SPEX is less than 0.1 seconds, which is negligible compared to the reasoning model's overall processing time. This demonstrates that SPEX's design introduces only minor latency overhead for reward evaluation, ensuring it does not become a bottleneck in the reasoning pipeline.

\subsection{Prediction Accuracy}

\begin{figure}
    \centering
    \includegraphics[width=\linewidth]{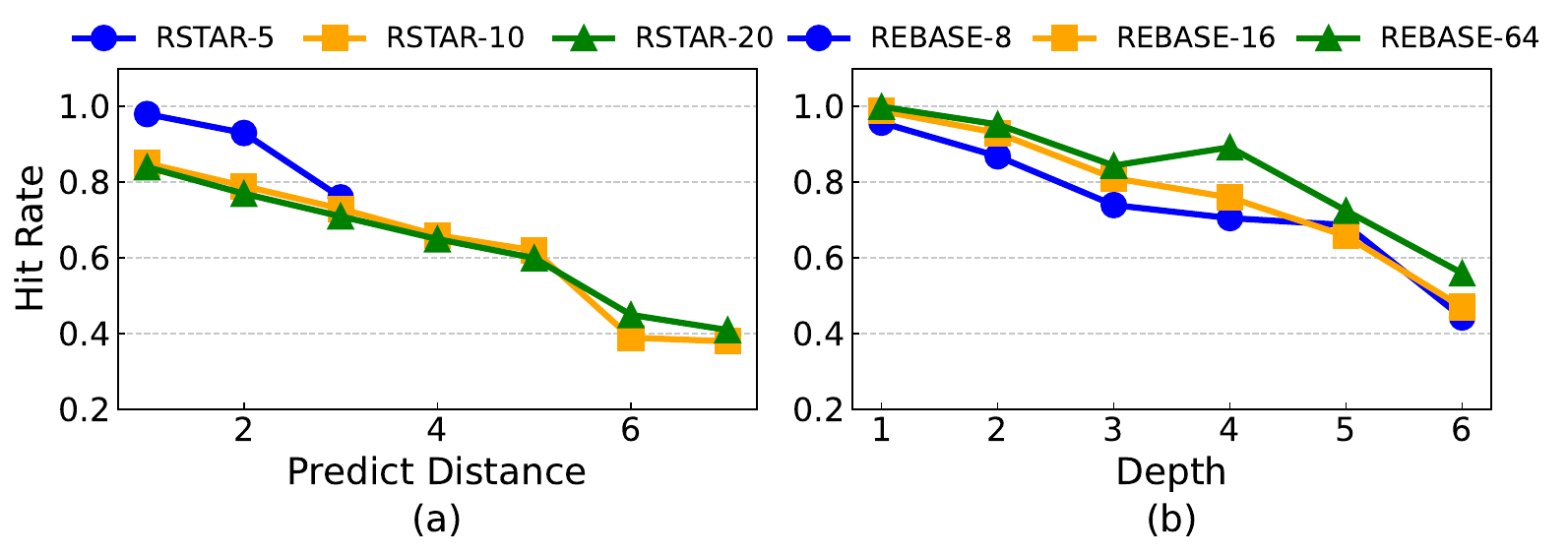}
    \caption{Prediction accuracy evaluation. (a) Hit rate of RSTAR-MCTS for predicting rollouts across different distances. (b) Hit rate of REBASE at varying depths.}
    \label{fig:specacc}
\end{figure}

Figure~\ref{fig:specacc} evaluates the prediction accuracy of SPEX. In (a), the hit rate of RSTAR-MCTS decreases as the prediction distance increases, illustrating the challenge of maintaining accuracy in DFS algorithms when predicting further rollouts. In (b), the hit rate of REBASE decreases with increasing depth, as shallow layers contain more low-scoring nodes that are easier to identify, while deeper layers consist of nodes with closer scores, making differentiation more difficult. Therefore, we prefer to allocate SPEX resources evenly for DFS algorithms and prioritize shallow layers for BFS algorithms.

\begin{figure}
    \centering
    \includegraphics[width=\linewidth]{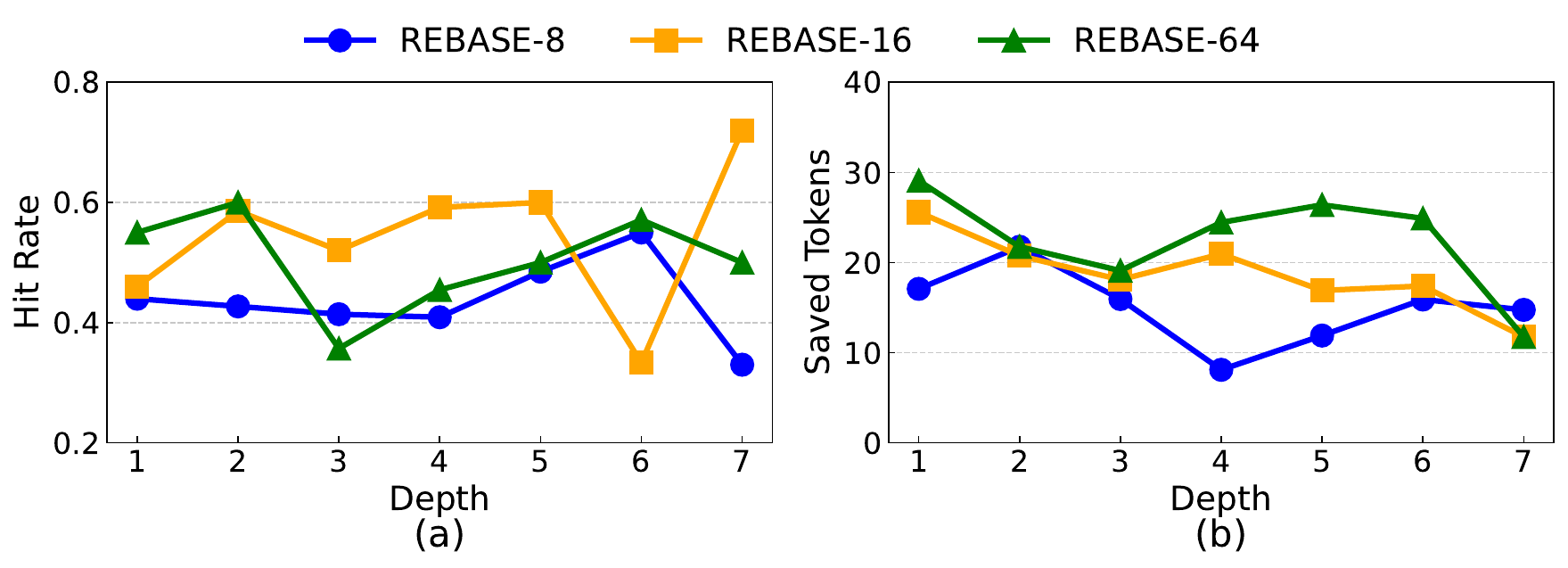}
    \caption{(a) Probability of speculative explorations reaching the critical path for REBASE configurations. (b) Tokens saved by speculation on the critical path.}
    \label{fig:critical}
\end{figure}

For DFS algorithms, prediction accuracy directly translates into speedup. However, for BFS-only algorithms like REBASE, only speculative expansions along the critical path lead to acceleration. As shown in Figure~\ref{fig:critical}(a), the probability of speculative expansions reaching the critical path ranges from 40\% to 60\% across different depths. In Figure~\ref{fig:critical}(b), the average number of tokens saved by aligning speculation with the critical path is approximately 20 per depth, highlighting the importance of accurate targeting for efficient performance.

\section{Related Work}

We categorize related work into three lines and position SPEX with respect to each.

\textbf{Speculative Decoding.}
Speculative decoding is an effective method for accelerating memory-bound LLM inference. It uses a smaller draft model to predict future tokens, which are then verified by the main model~\cite{xia2024unlocking,leviathan2023fast,chen2023accelerating}. Recent advancements have extended this technique to accelerate reasoning models~\cite{pan2025specreason,shispeccot,yang2025speculative,liao2025reward}. 
These techniques operate along a \emph{single linear trajectory}, speculating on the next-token sequence of one reasoning path. In contrast, SPEX speculates over the \emph{tree structure} of reasoning, predicting which branches will be needed before reward feedback is available. As we show in our evaluation, tree-level speculative exploration in SPEX is complementary to token-level speculative decoding and can be composed with it to yield cumulative speedups.

\textbf{Speculative Tree Search and Speculative Execution.}
Speculation has also been applied to Monte Carlo Tree Search (MCTS) and other search procedures. Prior work on speculative MCTS~\cite{NEURIPS2024_a19940b0,9576093} targets AlphaZero-style game search, where the bottleneck is the cost of reward evaluation. These systems speculate on future rollouts to amortize network computation, assuming a compute-bound regime and relatively cheap tree traversal. Our setting is fundamentally different: ToT reasoning with LLMs is predominantly memory-bandwidth-bound due to autoregressive decoding and KV-cache access. SPEX is designed specifically for this regime: it breaks the reward dependency barrier for both Depth-First and Breadth-First Search, incorporates KV-cache reuse and parameter reuse into its scheduling decisions, and coordinates speculation across concurrent queries. More broadly, hardware-level speculative execution mechanisms (e.g., branch prediction, value prediction, and speculation for FSM parallelization~\cite{3445814.3446705}) motivate using prediction to unlock latent parallelism; SPEX instantiates this principle at the ToT algorithm and system level for LLM reasoning workloads.

\textbf{Efficient Reasoning Algorithms.}
To improve reasoning efficiency, various algorithmic approaches focus on reducing the number of generated tokens. These methods include early exit \cite{yue2025don,fu2024efficiently,li2024escape,chen2024not}, CoT compression \cite{lin2025trimr,chen2023mcc,chen2025skip,zhao2025can,wang2025r1,hou2025thinkprune}, and adaptive reasoning \cite{liu2025can,qi2025optimizing,luo2025autol2s,wan2025adapthink,wang2025adaptive,lu2025prolonged}, which dynamically shorten or adjust reasoning paths based on task complexity or model confidence. These methods primarily optimize the \emph{amount} of test-time compute. SPEX is orthogonal: given a target ToT algorithm and its compute budget, we focus on \emph{how} to schedule and speculate over tree expansions to break the reward dependency barrier and improve hardware efficiency without changing the underlying reasoning algorithm.

\section{Conclusion}
 We propose SPEX, a speculative exploration framework that breaks the reward dependency barrier in Tree-of-Thought reasoning. By introducing intra-query speculative branch selection, inter-query budget allocation, and adaptive early termination, SPEX significantly enhances parallelism and reduces latency. Comprehensive evaluations demonstrate up to 3$\times$ speedup for DFS and 1.7$\times$ speedup for BFS, all while preserving accuracy. SPEX provides an effective and scalable solution for optimizing ToT reasoning, bridging the gap between algorithmic innovation and system-level performance.

\bibliographystyle{plain}
\bibliography{sample-base}

@article{hooper2025ets,
  title={Ets: Efficient tree search for inference-time scaling},
  author={Hooper, Coleman and Kim, Sehoon and Moon, Suhong and Dilmen, Kerem and Maheswaran, Monishwaran and Lee, Nicholas and Mahoney, Michael W and Shao, Sophia and Keutzer, Kurt and Gholami, Amir},
  journal={arXiv preprint arXiv:2502.13575},
  year={2025}
}

@inproceedings{NEURIPS2024_a19940b0,
 author = {Cheng, Scott and Kandemir, Mahmut Taylan and Hong, Ding-Yong},
 booktitle = {Advances in Neural Information Processing Systems},
 doi = {10.52202/079017-2813},
 pages = {88664--88683},
 title = {Speculative Monte-Carlo Tree Search},
 year = {2024}
}

@ARTICLE{9576093,
  author={Kim, Juhwan and Kang, Byeongmin and Cho, Hyungmin},
  journal={IEEE Access}, 
  title={SpecMCTS: Accelerating Monte Carlo Tree Search Using Speculative Tree Traversal}, 
  year={2021},
  volume={9},
  number={},
  pages={142195-142205}
}

@inproceedings{3445814.3446705,
author = {Qiu, Junqiao and Sun, Xiaofan and Sabet, Amir Hossein Nodehi and Zhao, Zhijia},
title = {Scalable FSM parallelization via path fusion and higher-order speculation},
year = {2021},
isbn = {9781450383172},
publisher = {Association for Computing Machinery},
address = {New York, NY, USA},
url = {https://doi.org/10.1145/3445814.3446705},
doi = {10.1145/3445814.3446705},
booktitle = {Proceedings of the 26th ACM International Conference on Architectural Support for Programming Languages and Operating Systems},
pages = {887–901},
numpages = {15},
location = {Virtual, USA},
series = {ASPLOS '21}
}

@inproceedings{wu2025inference,
  title={Inference scaling laws: An empirical analysis of compute-optimal inference for LLM problem-solving},
  author={Wu, Yangzhen and Sun, Zhiqing and Li, Shanda and Welleck, Sean and Yang, Yiming},
  booktitle={The Thirteenth International Conference on Learning Representations},
  year={2025}
}

@article{chu2023navigate,
  title={Navigate through enigmatic labyrinth a survey of chain of thought reasoning: Advances, frontiers and future},
  author={Chu, Zheng and Chen, Jingchang and Chen, Qianglong and Yu, Weijiang and He, Tao and Wang, Haotian and Peng, Weihua and Liu, Ming and Qin, Bing and Liu, Ting},
  journal={arXiv preprint arXiv:2309.15402},
  year={2023}
}

@article{yao2023tree,
  title={Tree of thoughts: Deliberate problem solving with large language models},
  author={Yao, Shunyu and Yu, Dian and Zhao, Jeffrey and Shafran, Izhak and Griffiths, Tom and Cao, Yuan and Narasimhan, Karthik},
  journal={Advances in neural information processing systems},
  volume={36},
  pages={11809--11822},
  year={2023}
}

@article{wei2022chain,
  title={Chain-of-thought prompting elicits reasoning in large language models},
  author={Wei, Jason and Wang, Xuezhi and Schuurmans, Dale and Bosma, Maarten and Xia, Fei and Chi, Ed and Le, Quoc V and Zhou, Denny and others},
  journal={Advances in neural information processing systems},
  volume={35},
  pages={24824--24837},
  year={2022}
}

@article{wang2024openr,
  title={Openr: An open source framework for advanced reasoning with large language models},
  author={Wang, Jun and Fang, Meng and Wan, Ziyu and Wen, Muning and Zhu, Jiachen and Liu, Anjie and Gong, Ziqin and Song, Yan and Chen, Lei and Ni, Lionel M and others},
  journal={arXiv preprint arXiv:2410.09671},
  year={2024}
}

@article{zhang2024rest,
  title={Rest-mcts*: Llm self-training via process reward guided tree search},
  author={Zhang, Dan and Zhoubian, Sining and Hu, Ziniu and Yue, Yisong and Dong, Yuxiao and Tang, Jie},
  journal={Advances in Neural Information Processing Systems},
  volume={37},
  pages={64735--64772},
  year={2024}
}

@article{qi2024mutual,
  title={Mutual reasoning makes smaller llms stronger problem-solvers},
  author={Qi, Zhenting and Ma, Mingyuan and Xu, Jiahang and Zhang, Li Lyna and Yang, Fan and Yang, Mao},
  journal={arXiv preprint arXiv:2408.06195},
  year={2024}
}

@inproceedings{kwon2023efficient,
  title={Efficient memory management for large language model serving with pagedattention},
  author={Kwon, Woosuk and Li, Zhuohan and Zhuang, Siyuan and Sheng, Ying and Zheng, Lianmin and Yu, Cody Hao and Gonzalez, Joseph and Zhang, Hao and Stoica, Ion},
  booktitle={Proceedings of the 29th symposium on operating systems principles},
  pages={611--626},
  year={2023}
}

@article{snell2024scaling,
  title={Scaling llm test-time compute optimally can be more effective than scaling model parameters},
  author={Snell, Charlie and Lee, Jaehoon and Xu, Kelvin and Kumar, Aviral},
  journal={arXiv preprint arXiv:2408.03314},
  year={2024}
}

@article{zheng2024sglang,
  title={Sglang: Efficient execution of structured language model programs},
  author={Zheng, Lianmin and Yin, Liangsheng and Xie, Zhiqiang and Sun, Chuyue Livia and Huang, Jeff and Yu, Cody Hao and Cao, Shiyi and Kozyrakis, Christos and Stoica, Ion and Gonzalez, Joseph E and others},
  journal={Advances in neural information processing systems},
  volume={37},
  pages={62557--62583},
  year={2024}
}

@article{fu2024efficiently,
  title={Efficiently Scaling LLM Reasoning with Certaindex},
  author={Fu, Yichao and Chen, Junda and Zhu, Siqi and Fu, Zheyu and Dai, Zhongdongming and Zhuang, Yonghao and Ma, Yian and Qiao, Aurick and Rosing, Tajana and Stoica, Ion and others},
  journal={arXiv preprint arXiv:2412.20993},
  year={2024}
}

@article{chen2023mcc,
  title={Mcc-kd: Multi-cot consistent knowledge distillation},
  author={Chen, Hongzhan and Wu, Siyue and Quan, Xiaojun and Wang, Rui and Yan, Ming and Zhang, Ji},
  journal={arXiv preprint arXiv:2310.14747},
  year={2023}
}

@article{liu2025can,
  title={Can 1b llm surpass 405b llm? rethinking compute-optimal test-time scaling},
  author={Liu, Runze and Gao, Junqi and Zhao, Jian and Zhang, Kaiyan and Li, Xiu and Qi, Biqing and Ouyang, Wanli and Zhou, Bowen},
  journal={arXiv preprint arXiv:2502.06703},
  year={2025}
}

@article{chen2025skip,
  title={Skip-thinking: Chunk-wise chain-of-thought distillation enable smaller language models to reason better and faster},
  author={Chen, Xiao and Zhou, Sihang and Liang, Ke and Sun, Xiaoyu and Liu, Xinwang},
  journal={arXiv preprint arXiv:2505.18642},
  year={2025}
}

@article{zhao2025can,
  title={Can pruning improve reasoning? revisiting long-cot compression with capability in mind for better reasoning},
  author={Zhao, Shangziqi and Yuan, Jiahao and Yang, Guisong and Naseem, Usman},
  journal={arXiv preprint arXiv:2505.14582},
  year={2025}
}

@article{wang2025r1,
  title={R1-Compress: Long Chain-of-Thought Compression via Chunk Compression and Search},
  author={Wang, Yibo and Shen, Li and Yao, Huanjin and Huang, Tiansheng and Liu, Rui and Tan, Naiqiang and Huang, Jiaxing and Zhang, Kai and Tao, Dacheng},
  journal={arXiv preprint arXiv:2505.16838},
  year={2025}
}

@article{lin2025trimr,
  title={TrimR: Verifier-based Training-Free Thinking Compression for Efficient Test-Time Scaling},
  author={Lin, Weizhe and Li, Xing and Yang, Zhiyuan and Fu, Xiaojin and Zhen, Hui-Ling and Wang, Yaoyuan and Yu, Xianzhi and Liu, Wulong and Li, Xiaosong and Yuan, Mingxuan},
  journal={arXiv preprint arXiv:2505.17155},
  year={2025}
}

@article{beeching2024scaling,
  title = {Scaling Test-Time Compute with Open Models},
  author = {Beeching, Edward and Tunstall, Lewis and Rush, Sasha},
  year = {2024},
  url = {https://huggingface.co/spaces/HuggingFaceH4/blogpost-scaling-test-time-compute},
  note = {Hugging Face Technical Report}
}

@article{kojima2022large,
  title={Large language models are zero-shot reasoners},
  author={Kojima, Takeshi and Gu, Shixiang Shane and Reid, Machel and Matsuo, Yutaka and Iwasawa, Yusuke},
  journal={Advances in neural information processing systems},
  volume={35},
  pages={22199--22213},
  year={2022}
}

@article{zou2023generalizable,
  title={Generalizable chain-of-thought prompting in mixed-task scenarios with large language models},
  author={Zou, Anni and Zhang, Zhuosheng and Zhao, Hai and Tang, Xiangru},
  journal={arXiv preprint arXiv:2310.06692},
  year={2023}
}

@inproceedings{chaslot2008monte,
  title={Monte-carlo tree search: A new framework for game ai},
  author={Chaslot, Guillaume and Bakkes, Sander and Szita, Istvan and Spronck, Pieter},
  booktitle={Proceedings of the AAAI Conference on Artificial Intelligence and Interactive Digital Entertainment},
  volume={4},
  number={1},
  pages={216--217},
  year={2008}
}

@article{xie2024monte,
  title={Monte carlo tree search boosts reasoning via iterative preference learning},
  author={Xie, Yuxi and Goyal, Anirudh and Zheng, Wenyue and Kan, Min-Yen and Lillicrap, Timothy P and Kawaguchi, Kenji and Shieh, Michael},
  journal={arXiv preprint arXiv:2405.00451},
  year={2024}
}

@article{cheng2024self,
  title={Self-playing adversarial language game enhances llm reasoning},
  author={Cheng, Pengyu and Dai, Yong and Hu, Tianhao and Xu, Han and Zhang, Zhisong and Han, Lei and Du, Nan and Li, Xiaolong},
  journal={Advances in Neural Information Processing Systems},
  volume={37},
  pages={126515--126543},
  year={2024}
}

@article{ning2023skeleton,
  title={Skeleton-of-thought: Prompting llms for efficient parallel generation},
  author={Ning, Xuefei and Lin, Zinan and Zhou, Zixuan and Wang, Zifu and Yang, Huazhong and Wang, Yu},
  journal={arXiv preprint arXiv:2307.15337},
  year={2023}
}

@techreport{openai2025o3mini,
  title = {{OpenAI O3-mini System Card}},
  author = {{OpenAI}},
  institution = {OpenAI},
  year = {2025},
  month = {January},
  note = {Technical Report}
}

@article{R1,
  title={Deepseek-r1: Incentivizing reasoning capability in llms via reinforcement learning},
  author={Guo, Daya and Yang, Dejian and Zhang, Haowei and Song, Junxiao and Zhang, Ruoyu and Xu, Runxin and Zhu, Qihao and Ma, Shirong and Wang, Peiyi and Bi, Xiao and others},
  journal={arXiv preprint arXiv:2501.12948},
  year={2025}
}

@techreport{openai2024learning,
  title = {Learning to Reason with LLMs},
  author = {OpenAI},
  year = {2024},
  url = {https://openai.com/index/learning-to-reason-with-llms/}, 
  note = {Technical Report},
  institution = {OpenAI}
}

@article{cobbe2021training,
  title={Training verifiers to solve math word problems},
  author={Cobbe, Karl and Kosaraju, Vineet and Bavarian, Mohammad and Chen, Mark and Jun, Heewoo and Kaiser, Lukasz and Plappert, Matthias and Tworek, Jerry and Hilton, Jacob and Nakano, Reiichiro and others},
  journal={arXiv preprint arXiv:2110.14168},
  year={2021}
}

@article{hendrycks2021measuring,
  title={Measuring mathematical problem solving with the math dataset},
  author={Hendrycks, Dan and Burns, Collin and Kadavath, Saurav and Arora, Akul and Basart, Steven and Tang, Eric and Song, Dawn and Steinhardt, Jacob},
  journal={arXiv preprint arXiv:2103.03874},
  year={2021}
}

@article{azerbayev2023llemma,
  title={Llemma: An open language model for mathematics},
  author={Azerbayev, Zhangir and Schoelkopf, Hailey and Paster, Keiran and Santos, Marco Dos and McAleer, Stephen and Jiang, Albert Q and Deng, Jia and Biderman, Stella and Welleck, Sean},
  journal={arXiv preprint arXiv:2310.10631},
  year={2023}
}

@inproceedings{lightman2023let,
  title={Let's verify step by step},
  author={Lightman, Hunter and Kosaraju, Vineet and Burda, Yuri and Edwards, Harrison and Baker, Bowen and Lee, Teddy and Leike, Jan and Schulman, John and Sutskever, Ilya and Cobbe, Karl},
  booktitle={The Twelfth International Conference on Learning Representations},
  year={2023}
}

@article{qiu2024treebon,
  title={Treebon: Enhancing inference-time alignment with speculative tree-search and best-of-n sampling},
  author={Qiu, Jiahao and Lu, Yifu and Zeng, Yifan and Guo, Jiacheng and Geng, Jiayi and Wang, Huazheng and Huang, Kaixuan and Wu, Yue and Wang, Mengdi},
  journal={arXiv preprint arXiv:2410.16033},
  year={2024}
}

@article{gao2024interpretable,
  title={Interpretable contrastive monte carlo tree search reasoning},
  author={Gao, Zitian and Niu, Boye and He, Xuzheng and Xu, Haotian and Liu, Hongzhang and Liu, Aiwei and Hu, Xuming and Wen, Lijie},
  journal={arXiv preprint arXiv:2410.01707},
  year={2024}
}

@article{yuan2024advancing,
  title={Advancing llm reasoning generalists with preference trees},
  author={Yuan, Lifan and Cui, Ganqu and Wang, Hanbin and Ding, Ning and Wang, Xingyao and Deng, Jia and Shan, Boji and Chen, Huimin and Xie, Ruobing and Lin, Yankai and others},
  journal={arXiv preprint arXiv:2404.02078},
  year={2024}
}

@misc{team2025sky,
  title={Sky-t1: Train your own o1 preview model within \$450},
  author={Team, NovaSky},
  year={2025}
}

@article{team2024qwq,
  title={Qwq: Reflect deeply on the boundaries of the unknown},
  author={Team, Qwen},
  journal={Hugging Face},
  year={2024}
}

@article{brown2024large,
  title={Large language monkeys: Scaling inference compute with repeated sampling},
  author={Brown, Bradley and Juravsky, Jordan and Ehrlich, Ryan and Clark, Ronald and Le, Quoc V and R{\'e}, Christopher and Mirhoseini, Azalia},
  journal={arXiv preprint arXiv:2407.21787},
  year={2024}
}

@article{chen2024more,
  title={Are more llm calls all you need? towards the scaling properties of compound ai systems},
  author={Chen, Lingjiao and Davis, Jared Quincy and Hanin, Boris and Bailis, Peter and Stoica, Ion and Zaharia, Matei A and Zou, James Y},
  journal={Advances in Neural Information Processing Systems},
  volume={37},
  pages={45767--45790},
  year={2024}
}

@article{zhangmore,
  title={More Agents Is All You Need},
  author={Zhang, Qin and Yu, Yangbin and FU, QIANG and Ye, Deheng and others},
  journal={Transactions on Machine Learning Research}
}

@inproceedings{li2023making,
  title={Making language models better reasoners with step-aware verifier},
  author={Li, Yifei and Lin, Zeqi and Zhang, Shizhuo and Fu, Qiang and Chen, Bei and Lou, Jian-Guang and Chen, Weizhu},
  booktitle={Proceedings of the 61st Annual Meeting of the Association for Computational Linguistics (Volume 1: Long Papers)},
  pages={5315--5333},
  year={2023}
}

@article{feng2023alphazero,
  title={Alphazero-like tree-search can guide large language model decoding and training},
  author={Feng, Xidong and Wan, Ziyu and Wen, Muning and McAleer, Stephen Marcus and Wen, Ying and Zhang, Weinan and Wang, Jun},
  journal={arXiv preprint arXiv:2309.17179},
  year={2023}
}

@article{wang2024q,
  title={Q*: Improving multi-step reasoning for llms with deliberative planning},
  author={Wang, Chaojie and Deng, Yanchen and Lyu, Zhiyi and Zeng, Liang and He, Jujie and Yan, Shuicheng and An, Bo},
  journal={arXiv preprint arXiv:2406.14283},
  year={2024}
}

@article{hou2025thinkprune,
  title={Thinkprune: Pruning long chain-of-thought of llms via reinforcement learning},
  author={Hou, Bairu and Zhang, Yang and Ji, Jiabao and Liu, Yujian and Qian, Kaizhi and Andreas, Jacob and Chang, Shiyu},
  journal={arXiv preprint arXiv:2504.01296},
  year={2025}
}

@article{chen2024not,
  title={Do not think that much for 2+ 3=? on the overthinking of o1-like llms},
  author={Chen, Xingyu and Xu, Jiahao and Liang, Tian and He, Zhiwei and Pang, Jianhui and Yu, Dian and Song, Linfeng and Liu, Qiuzhi and Zhou, Mengfei and Zhang, Zhuosheng and others},
  journal={arXiv preprint arXiv:2412.21187},
  year={2024}
}

@article{qi2025optimizing,
  title={Optimizing anytime reasoning via budget relative policy optimization},
  author={Qi, Penghui and Liu, Zichen and Pang, Tianyu and Du, Chao and Lee, Wee Sun and Lin, Min},
  journal={arXiv preprint arXiv:2505.13438},
  year={2025}
}

@article{li2024large,
  title={Large language model inference acceleration: A comprehensive hardware perspective},
  author={Li, Jinhao and Xu, Jiaming and Huang, Shan and Chen, Yonghua and Li, Wen and Liu, Jun and Lian, Yaoxiu and Pan, Jiayi and Ding, Li and Zhou, Hao and others},
  journal={arXiv preprint arXiv:2410.04466},
  year={2024}
}

@article{li2024escape,
  title={Escape sky-high cost: Early-stopping self-consistency for multi-step reasoning},
  author={Li, Yiwei and Yuan, Peiwen and Feng, Shaoxiong and Pan, Boyuan and Wang, Xinglin and Sun, Bin and Wang, Heda and Li, Kan},
  journal={arXiv preprint arXiv:2401.10480},
  year={2024}
}

@article{xia2024unlocking,
  title={Unlocking efficiency in large language model inference: A comprehensive survey of speculative decoding},
  author={Xia, Heming and Yang, Zhe and Dong, Qingxiu and Wang, Peiyi and Li, Yongqi and Ge, Tao and Liu, Tianyu and Li, Wenjie and Sui, Zhifang},
  journal={arXiv preprint arXiv:2401.07851},
  year={2024}
}

@inproceedings{leviathan2023fast,
  title={Fast inference from transformers via speculative decoding},
  author={Leviathan, Yaniv and Kalman, Matan and Matias, Yossi},
  booktitle={International Conference on Machine Learning},
  pages={19274--19286},
  year={2023},
  organization={PMLR}
}

@article{chen2023accelerating,
  title={Accelerating large language model decoding with speculative sampling},
  author={Chen, Charlie and Borgeaud, Sebastian and Irving, Geoffrey and Lespiau, Jean-Baptiste and Sifre, Laurent and Jumper, John},
  journal={arXiv preprint arXiv:2302.01318},
  year={2023}
}

@article{pan2025specreason,
  title={Specreason: Fast and accurate inference-time compute via speculative reasoning},
  author={Pan, Rui and Dai, Yinwei and Zhang, Zhihao and Oliaro, Gabriele and Jia, Zhihao and Netravali, Ravi},
  journal={arXiv preprint arXiv:2504.07891},
  year={2025}
}

@inproceedings{shispeccot,
  title={SpecCoT: Accelerating Chain-of-Thought Reasoning through Speculative Exploration},
  author={Shi, Junhan and Zhu, Yijia and Shi, Zhenning and Zhao, Dan and Li, Qing and Jiang, Yong},
  booktitle={ES-FoMo III: 3rd Workshop on Efficient Systems for Foundation Models}
}

@article{yang2025speculative,
  title={Speculative thinking: Enhancing small-model reasoning with large model guidance at inference time},
  author={Yang, Wang and Yue, Xiang and Chaudhary, Vipin and Han, Xiaotian},
  journal={arXiv preprint arXiv:2504.12329},
  year={2025}
}

@article{liao2025reward,
  title={Reward-guided speculative decoding for efficient llm reasoning},
  author={Liao, Baohao and Xu, Yuhui and Dong, Hanze and Li, Junnan and Monz, Christof and Savarese, Silvio and Sahoo, Doyen and Xiong, Caiming},
  journal={arXiv preprint arXiv:2501.19324},
  year={2025}
}

@article{yue2025don,
  title={Don't Overthink It: A Survey of Efficient R1-style Large Reasoning Models},
  author={Yue, Linan and Du, Yichao and Wang, Yizhi and Gao, Weibo and Yao, Fangzhou and Wang, Li and Liu, Ye and Xu, Ziyu and Liu, Qi and Di, Shimin and others},
  journal={arXiv preprint arXiv:2508.02120},
  year={2025}
}

@article{ye2025flashinfer,
  title={Flashinfer: Efficient and customizable attention engine for llm inference serving},
  author={Ye, Zihao and Chen, Lequn and Lai, Ruihang and Lin, Wuwei and Zhang, Yineng and Wang, Stephanie and Chen, Tianqi and Kasikci, Baris and Grover, Vinod and Krishnamurthy, Arvind and others},
  journal={arXiv preprint arXiv:2501.01005},
  year={2025}
}

@article{dao2022flashattention,
  title={Flashattention: Fast and memory-efficient exact attention with io-awareness},
  author={Dao, Tri and Fu, Dan and Ermon, Stefano and Rudra, Atri and R{\'e}, Christopher},
  journal={Advances in neural information processing systems},
  volume={35},
  pages={16344--16359},
  year={2022}
}

@article{luo2025autol2s,
  title={Autol2s: Auto long-short reasoning for efficient large language models},
  author={Luo, Feng and Chuang, Yu-Neng and Wang, Guanchu and Le, Hoang Anh Duy and Zhong, Shaochen and Liu, Hongyi and Yuan, Jiayi and Sui, Yang and Braverman, Vladimir and Chaudhary, Vipin and others},
  journal={arXiv preprint arXiv:2505.22662},
  year={2025}
}

@article{wan2025adapthink,
  title={AdapThink: Adaptive Thinking Preferences for Reasoning Language Model},
  author={Wan, Xu and Wang, Wei and Xu, Wenyue and Yin, Wotao and Song, Jie and Sun, Mingyang},
  journal={arXiv preprint arXiv:2506.18237},
  year={2025}
}

@article{lu2025prolonged,
  title={Prolonged reasoning is not all you need: Certainty-based adaptive routing for efficient llm/mllm reasoning},
  author={Lu, Jinghui and Yu, Haiyang and Xu, Siliang and Ran, Shiwei and Tang, Guozhi and Wang, Siqi and Shan, Bin and Fu, Teng and Feng, Hao and Tang, Jingqun and others},
  journal={arXiv preprint arXiv:2505.15154},
  year={2025}
}

@article{wang2025adaptive,
  title={Adaptive Deep Reasoning: Triggering Deep Thinking When Needed},
  author={Wang, Yunhao and Zhang, Yuhao and Yu, Tinghao and Xu, Can and Zhang, Feng and Lian, Fengzong},
  journal={arXiv preprint arXiv:2505.20101},
  year={2025}
}

@misc{fu2025deepthinkconfidence,
      title={Deep Think with Confidence}, 
      author={Yichao Fu and Xuewei Wang and Yuandong Tian and Jiawei Zhao},
      year={2025},
      eprint={2508.15260},
      archivePrefix={arXiv},
      primaryClass={cs.LG},
      url={https://arxiv.org/abs/2508.15260}, 
}

@misc{aime2024,
  title        = {{American Invitational Mathematics Examination (AIME) 2024}},
  author       = {{Mathematical Association of America}},
  year         = {2024},
  howpublished = {\url{https://www.maa.org/math-competitions/aime}}
}

@misc{aime2025,
  title        = {{American Invitational Mathematics Examination (AIME) 2025}},
  author       = {{Mathematical Association of America}},
  year         = {2025},
  howpublished = {\url{https://www.maa.org/math-competitions/aime}}
}

@misc{hmmt2025,
  title        = {{Harvard-MIT Mathematics Tournament (HMMT) February 2025}},
  author       = {{HMMT Organization}},
  year         = {2025},
  howpublished = {\url{https://www.hmmt.org}}
}

@misc{brumo2025,
  title        = {{brown university math olympiad 2025}},
  author       = {{BRUMO}},
  year         = {2025},
  howpublished = {\url{https://www.brumo.org/}}
}

@misc{yang2025qwen3technicalreport,
      title={Qwen3 Technical Report}, 
      author={An Yang and Anfeng Li and Baosong Yang and Beichen Zhang and Binyuan Hui and Bo Zheng and Bowen Yu and Chang Gao and Chengen Huang and Chenxu Lv and Chujie Zheng and Dayiheng Liu and Fan Zhou and Fei Huang and Feng Hu and Hao Ge and Haoran Wei and Huan Lin and Jialong Tang and Jian Yang and Jianhong Tu and Jianwei Zhang and Jianxin Yang and Jiaxi Yang and Jing Zhou and Jingren Zhou and Junyang Lin and Kai Dang and Keqin Bao and Kexin Yang and Le Yu and Lianghao Deng and Mei Li and Mingfeng Xue and Mingze Li and Pei Zhang and Peng Wang and Qin Zhu and Rui Men and Ruize Gao and Shixuan Liu and Shuang Luo and Tianhao Li and Tianyi Tang and Wenbiao Yin and Xingzhang Ren and Xinyu Wang and Xinyu Zhang and Xuancheng Ren and Yang Fan and Yang Su and Yichang Zhang and Yinger Zhang and Yu Wan and Yuqiong Liu and Zekun Wang and Zeyu Cui and Zhenru Zhang and Zhipeng Zhou and Zihan Qiu},
      year={2025},
      eprint={2505.09388},
      archivePrefix={arXiv},
      primaryClass={cs.CL},
      url={https://arxiv.org/abs/2505.09388}, 
}

@inproceedings{orches,
author = {Li, Sixu and Chen, Yuzhou and Li, Chaojian and Fu, Yonggan and Wang, Zheng and Yu, Zhongzhi and You, Haoran and Ye, Zhifan and Zhou, Wei and Zhang, Yongan and Lin, Yingyan (Celine)},
title = {ORCHES: Orchestrated Test-Time-Compute-based LLM Reasoning on Collaborative GPU-PIM HEterogeneous System},
year = {2025},
isbn = {9798400715730},
publisher = {Association for Computing Machinery},
address = {New York, NY, USA},
url = {https://doi.org/10.1145/3725843.3756039},
doi = {10.1145/3725843.3756039},
booktitle = {Proceedings of the 58th IEEE/ACM International Symposium on Microarchitecture},
pages = {476–489},
numpages = {14},
location = {
},
series = {MICRO '25}
}

\end{document}